\documentclass{article} %
\usepackage{iclr2027_conference,times}

\usepackage{amsmath,amsfonts,bm}

\def\eqref#1{equation~\ref{#1}}

\def\1{\bm{1}}

\DeclareMathAlphabet{\mathsfit}{\encodingdefault}{\sfdefault}{m}{sl}
\SetMathAlphabet{\mathsfit}{bold}{\encodingdefault}{\sfdefault}{bx}{n}

\usepackage{footmisc} %
\usepackage{hyperref}
\usepackage{url}

\usepackage{amsmath}
\usepackage{amssymb}
\usepackage[short]{optidef}
\usepackage{booktabs}
\usepackage{tcolorbox}
\tcbuselibrary{breakable}
\usepackage{siunitx}
\usepackage{caption}
\usepackage{fvextra}
\usepackage{fancyvrb}
\usepackage{wrapfig}
\usepackage{makecell}
\usepackage{inconsolata}
\usepackage{comment}
\definecolor{reframed-green}{HTML}{236623}
\definecolor{reframed-silver}{HTML}{B8B8B7}
\definecolor{reframed-gray}{HTML}{CFCFCE}
\definecolor{reframed-yellow}{HTML}{E8B017}
\definecolor{reframed-gold}{HTML}{DA9100}
\definecolor{reframed-red}{HTML}{B93B32}

\usepackage{xcolor}
\usepackage{listings}
\usepackage{colortbl}
\usepackage[sort&compress]{cleveref}

\usepackage{tikz}
\usetikzlibrary{arrows.meta,patterns,backgrounds,calc,positioning}
\usepackage{pgfplots}
\pgfplotsset{compat=1.18}
\usepgfplotslibrary{groupplots}
\usetikzlibrary{arrows.meta}
\usetikzlibrary{pgfplots.statistics}
\usepackage[normalem]{ulem}
\usepackage{lscape}

\usepackage{enumitem}
\setlist[itemize]{leftmargin=1.5em}

\usepackage{bbm}

\newcommand{\citeposs}[1]{\citeauthor{#1}'s \citeyearpar{#1}}

\usepackage{array}

\title{What, When, and How: Audio Description\\as Constrained Global Optimization}

\author{Igor Sterner, Mirella Lapata, Alex Lascarides
\& Frank Keller \\
  School of Informatics \\
  University of Edinburgh \\
  United Kingdom \\
  {\tt igor.sterner@ed.ac.uk, }%
  \{%
  {\tt mlap,alex,keller}%
  \}%
  {\tt @inf.ed.ac.uk}
}

\iclrfinalcopy %
\begin{document}

\maketitle
\lhead{Preprint. Under review.}
\begin{abstract}
Audio Description (AD) makes movies accessible to blind and visually impaired audiences by narrating visual information in gaps between dialogue. Existing automatic AD systems largely treat generation as a local video-to-text problem, assuming that the content to describe and its temporal location are already provided. Realistic AD instead requires coupled decisions about \emph{what} visual information is narratively important, \emph{when} it can be spoken without interfering with dialogue, and \emph{how} it should be formulated to fit within the available time. We formalize AD generation as a constrained optimization problem over these three decisions. Our hybrid system uses large language models to propose and ground visual elements, estimate their salience to the narrative, and generate compressed realizations. A mixed-integer linear program then jointly selects and schedules descriptions across a scene subject to temporal constraints. When evaluated on REFRAMED, a benchmark for realistic AD of movies, our approach  makes better decisions than prompted LLMs about what to describe and when to describe it, establishing a new SOTA on narrative QA and temporally grounded metrics. Ablations show that explicit temporal constraints drive gains in placement, while salience estimation controls how much narratively useful content is retained. Improvements are concentrated on temporal and narrative measures rather than n-gram overlap, although a significant gap to professional describers remains.
\end{abstract}

\section{Introduction}

Audio Description (AD) is a verbal narration of the key visual content of a movie, delivered in the gaps between dialogue so that blind and visually impaired audiences can follow the story. 
The central challenge of AD generation is deciding what visual information is most relevant, and what information can be left out \citep{vercauteren-2016-narratological}.
A movie records countless details its director never deliberately chose (the colour of a passing car, the weather, what an extra is wearing), which is in contrast to a verbal narrative which contains only the details its narrator selected and asserted \citep{chatman-1980-novels}.
Describing a movie therefore means recovering the small number of visual details that the director intended an audience to read as story, and omitting the overwhelming majority that merely happen to be in frame.
\citet{vercauteren-2007-what_when_how} characterizes the describer's craft as a set of interacting decisions about the \emph{what}, \emph{when}, and \emph{how} of narration: which visual elements carry the narrative, at what moment they can be spoken, and in what words.

The need to automate this work is now acute. In the US, Title II of the Americans with Disabilities Act will require public entities to provide AD for pre-recorded video, and the UK's Media Act sets a streaming quota of 10\% by 2030, a volume that fully human description,
which is slow and costly, cannot meet alone. Computational work has so far addressed largely only the last of the three decisions, reducing the task to a form of video captioning: the temporal placement of each description is taken from the reference AD, a clip is cut around it, and a model is trained or prompted to caption that clip, optionally conditioned on character information, screenplay context, or professional guidelines \citep{han-2023-autoad_i,han-2023-autoad_ii,xie-2024-autoad_zero,park-2025-narr_ad,li-2025-videoa11y}. This is a reasonable proxy for \emph{how} a description should be phrased, and multimodal LLMs are now strong at describing an input video \citep{alayrac-2022-flamingo,liu-2023-llava,garg-2024-imageinwords,chai-2025-auroracap,qwenteam-2026-qwen35omni}. However, it bypasses the problem of \emph{what} to describe and \emph{when} to describe it, by providing these as input to the model.

\begin{figure}[t]
\centering
\begin{tcolorbox}[
  colback=reframed-gray!10,
  colframe=reframed-silver,
  boxrule=0.5pt,
  arc=3pt,
  left=5pt, right=5pt, top=5pt, bottom=5pt,
  width=\columnwidth,
  fontupper=\scriptsize
]

[2s] \textit{Lisbeth is in an upmarket tailors.}
[3s] \textit{The tailor shows her a leather motorcycle jacket he's made from a photo of Mikael wearing an identical garment.}
\smallskip

-- [10s] \textbf{[SALANDER] It's nice.} -- [12s] \textbf{[TAILOR] Your father?}\\
-- [13s] \textbf{[SALANDER] A friend.} -- [14s] \textbf{[TAILOR] Must be a very good friend.}
\smallskip

[15s] \textit{She looks surprised by the tailor's comment.}
\smallskip

[18s] \textit{Back at her apartment, she writes a card which has a picture of a horse on it}
[22s] \textit{and put it in an envelope.}
[32s] \textit{She addresses the envelope to ‘M’.}
\smallskip

[35s] \textit{Night.}
[35s] \textit{Lisbeth rides her motorcycle down a cobbled street with snow piled at the curbside.}
[45s] \textit{She parks on a corner outside Mikael's apartment}
[48s] \textit{and removes her crash helmet.}
[50s] \textit{She takes a package from the back of the bike,}
[52s] \textit{but stops dead when she sees Mikael leave his apartment with Erika.}

\smallskip
-- [58s] \textbf{[MIKAEL] We're late.} -- [60s] \textbf{[ERIKA] Fashionably late.}
\smallskip

[60s] \textit{She watches numbly as the couple walk away with their arms around each other.}
[78s] \textit{Lisbeth looks down as Mikael and Erika get into a waiting taxi.}
[87s] \textit{Lisbeth turns away and flings the package which has the card attached to it into a nearby dumpster.}
[93s] \textit{She puts her crash helmet back on,}
[95s] \textit{gets on her bike}
[96s] \textit{and starts it up.}
[97s] \textit{A solitary figure, she rides off into the night.}

\end{tcolorbox}
\vspace{-8pt}
\caption{Excerpt of Audio Description (\emph{italics}) and dialogue (\textbf{bold}) from \emph{The Girl with the Dragon Tattoo} (2011); timestamps mark the start of narration.}
\label{fig:dragon-tattoo}
\end{figure}

Figure~\ref{fig:dragon-tattoo} illustrates what is lost when AD is reduced to captioning a pre-selected video clip.
Consider  \emph{what} the describer selects from the closing scenes of \emph{The Girl with the Dragon Tattoo} (see video \href{https://www.rottentomatoes.com/m/the_girl_with_the_dragon_tattoo/videos/ofdvX951xl2S}{here}):
a leather jacket made for Mikael, a card addressed to ``M'', and, eighty seconds later, that  card attached to a package thrown into a dumpster. These details are selected because together they carry the narrative: Lisbeth has prepared a gift for Mikael, then discards it after seeing him leave with Erika,
while countless co-occurring visual details are omitted. Consider \emph{when} the descriptions are delivered: they occupy gaps in the dialogue and remain close to the visuals they describe, but need not be synchronous with them. The jacket, for example, is described before the exchange that refers to it, allowing the subsequent dialogue to be understood in context. Consider \emph{how} the content is formulated: descriptions are concise to fit the available gaps, while still conveying narrative meaning, from the laconic ``Night'' to the more evocative ``A solitary figure, she rides off into the night.''

These decisions cannot be made independently, because they compete for narration time. Dense dialogue leaves little space, forcing a choice between describing fewer elements, describing them more tersely, or displacing a description away from the moment it refers to, by up to ten seconds \citep{sterner-2026-reframed}. A natural solution is to learn a model that emits descriptions and timestamps jointly, but \citet{sterner-2026-reframed} show that LLMs asked to do so produce descriptions of the wrong length for the gap they are given, overlapping with dialogue, or far from the visuals they describe. These are hard numerical constraints and LLM decoding provides no natural mechanism for enforcing them.

We therefore make the control explicit: neural models describe what happens on screen and when, and interpret it, while decisions that couple events under hard constraints are left to a solver. Our approach proceeds in three stages. First, a multimodal LLM describes the scene, and the description is segmented into events, each grounded to the span of video in which it occurs. Second, each event is scored for \emph{narrative salience}: how much a viewer's understanding of the story would suffer were it omitted.
Each event's description is also compressed to several shorter lengths, so that the same content is available both as a full sentence and as a shorter variant. Third, a solver decides across the whole scene at once which events to describe, which verbalization of each to use, and when to deliver it, maximizing total salience subject to hard constraints: narration must fall within a dialogue gap, stay close to the event it describes, and not overlap other narration. Because selection is discrete while delivery time is continuous, this is a mixed-integer linear program (MILP), for which mature and highly efficient solvers exist.

We evaluate on the REFRAMED benchmark \citep{sterner-2026-reframed}, whose challenge set provides dual human-authored AD references and complementary measures of content, temporal alignment, and narrative comprehension. Across Qwen- and Gemini-based pipelines, constrained optimization yields its largest gains on the temporal and QA-based measures. The MILP also consistently outperforms an LLM scheduler given the same candidate events, occurrence spans, salience scores, narration durations, and dialogue gaps, demonstrating the value of explicit optimization. Our strongest system achieves the best automatic QA-based performance, although a substantial gap to human describers remains. Our contributions\footnote{We will release all code necessary to reproduce our experimental results.} are as follows:
\vspace{-5pt}
\begin{itemize}
\itemsep0pt
\item A formalization of realistic AD generation as a constrained  optimization problem that jointly decides \emph{what} to describe,  \emph{when} to describe it, and \emph{how} to formulate it, bringing  together decisions that prior computational approaches optimize only  partially or separately.

\item A hybrid LLM-optimization system in which LLMs describe, ground, score, and compress events, and a mixed-integer linear program selects what to narrate, in which form, and when.

\item Results on REFRAMED showing that constrained optimization outperforms an LLM making the same decisions from the same inputs, and conveys more of the story than prompted LLMs and captioning systems that are told when to describe.
\end{itemize}

\section{Related Work}\label{sec:related}

Most work on AD generation assumes that the temporal placement of each description is given by the reference AD.  This reduces the task to video captioning: a model describes  visually salient content of a pre-specified clip.  Most systems fine-tune multimodal models for this task, augmenting inputs with signals such as broader visual context or character information \citep{han-2023-autoad_i,han-2023-autoad_ii,han-2024-autoad_iii, lin-2024-movie_seq,deganutti-2025-dante, wang-2025-uni_ad,ye-2025-focusedad}, while others prompt LLMs zero-shot \citep{chu-2024-llm_ad,zhang-2024-mm_narator,xie-2024-autoad_zero}. Despite differences in architecture and conditioning information, these approaches  assume away the decision of \emph{when} to describe and, by fixing the clip from the reference AD, strongly constrain \emph{what} should be described.

A smaller body of work relaxes this assumption. \citet{pavel-2020-rescribe} use dynamic programming to place human-scripted descriptions within audio gaps, shortening the text or lengthening the source audio where needed. \citet{wang-2021-toward} predict insertion times from audiovisual inconsistency and select a description from a candidate set at each predicted time. \citet{gupta-2026-visual} jointly predict insertion times and grounded visual content, then generate a description for those visuals, while \citet{khandelwal-2025-coherent} optimize content for predetermined temporal slots.
These approaches address different subsets of the \emph{what}, \emph{when}, and \emph{how} decisions, but none optimizes all three jointly across a scene: \citet{pavel-2020-rescribe} assume human-written content; \citet{wang-2021-toward} and \citet{gupta-2026-visual} generate descriptions independently at predicted insertion points rather than allocating a shared narration budget; and \citet{khandelwal-2025-coherent} fix temporal placement.

\citet{sterner-2026-reframed} formulate realistic AD generation as jointly deciding \emph{what} to describe and \emph{when}, and introduce REFRAMED, with dual human-authored references and metrics for this setting. REFRAMED provides the task and evaluation framework, but leaves open how these coupled decisions should be made under the hard temporal constraints imposed by the soundtrack. We extend this formulation by making \emph{how}, the choice of wording and hence narration duration, an explicit decision, and optimize content selection, formulation, and placement jointly across the scene.
The \emph{what}/\emph{when}/\emph{how} decomposition itself comes from work on AD in translation studies \citep{vercauteren-2007-what_when_how}, where these decisions are guided by what the audience needs in order to follow the story \citep{vercauteren-2016-narratological}. 
These accounts are descriptive, however, and do not specify a computational mechanism for coordinating the three decisions.

Our formulation is closely related to global optimization approaches in text summarization, where salient content is selected under a limited linguistic budget. Early work casts summarization as global inference over relevance, redundancy, and length \citep{mcdonald-2007-study}, while integer-linear programming makes content selection explicit under a length constraint \citep{gillick-2008-icsi,gillick-2009-scalable}. Particularly relevant to our setting, such formulations can also choose among alternative compressed realizations of the same content \citep{madnani-2007-multiple,clarke-2008-global,martins-2009-summarization,berg-2011-jointly}, or jointly select and rewrite content under a shared budget \citep{woodsend-2010-automatic,woodsend-2012-multiple}. We adopt the same separation between models that score candidate content and a solver that chooses among candidates. AD, however, replaces a single length budget with irregular temporal windows fixed by the soundtrack, while also requiring the optimizer to decide where within those windows selected content should be delivered.

\section{Formalization}\label{sec:formalization}

Building on the decisions illustrated in Figure~\ref{fig:dragon-tattoo}, this section formalizes AD generation as  decisions and constraints over \emph{what} to say, \emph{when} to say it, and \emph{how} to say it. We first define at what level decisions are made and introduce terminology. We then introduce decision variables for selecting, placing, and compressing descriptions. Finally, we state AD generation as an optimization problem.

\textbf{Units and terminology.}  AD may describe events (something dynamic, such as Lisbeth parking), processes (something ongoing, such as Lisbeth riding her motorcycle), or states (something static, such as it being night-time).  Following \citet{bach-1986-eventuality}, we will refer to all of these as {\em eventualities}. Our formulation is as follows.

\begin{itemize}
\itemsep0pt
    \item Each scene depicts a number of eventualities \(e_i\in\mathcal{E}\), obtained by segmenting a prose scene description and indexed in textual order.
    \item Each eventuality \(e_i\) has a candidate description element \(w_i\).
    \item The eventuality is depicted in the video during an occurrence span $[\tau_i, \tau_i+\gamma_i]$.
    \item The available narration time may require a shorter formulation, so each element has $K+1$ variants $\mathcal{C}_i = \{c_{i0},\dots,c_{iK}\}$, where \(c_{i0}=w_i\) and $c_{ik}$ is its $k$-th compression.
    \item If described, the eventuality is assigned a delivery span $[d_i, d_i+L_i]$, where $L_i$ is the time required to narrate the variant used.
\end{itemize}

Occurrence and delivery spans are related but frequently non-equal: dialogue often prevents description at the moment an eventuality is on screen, in which case its description is delivered shortly before or after it.

\textbf{Decision variables and further notation.} There are three decision variables, reflecting the \emph{what}, \emph{when}, and
\emph{how} of AD.

\begin{itemize}
\itemsep0pt
    \item \emph{What} --- \(x_i \in \{ 0, 1 \}\) represents whether eventuality \(e_i\) is described or not.
    \item \emph{When} --- \(d_i \in \mathbb{R}_{\geq0}\) represents the delivery start time of the description of \(e_i\).
    \item \emph{How} --- \(y_{ik} \in \{0, 1\}\) represents whether variant \(c_{ik}\) is used.
\end{itemize}

Each eventuality is associated with a salience score \(\sigma_i\in \mathbb{R}\), which represents its importance to the story being told.
We define \(l_{ik} = \delta \left( c_{ik} \right)\), where \(\delta\) maps text to narration duration, so that the narration time of the variant used is \(L_i = \sum_{k=0}^{K} l_{ik} y_{ik}\).
The mid-point of the occurrence span of \(e_i\) is \(m_i = \tau_i + \frac{\gamma_i}{2}\).
Finally, $\mathcal{G} = \bigcup_{j=1}^{J} [a_j,b_j]$ is the set of permissible narration times in a scene, formed by the union of $J$ temporally disjoint closed intervals $[a_j,b_j]$.

\textbf{AD as an optimization problem.} Given candidate description elements and their variants, salience scores, occurrence spans, and permissible gaps, AD generation requires deciding which eventualities
to describe, which variant to use for each, and when to deliver it.
We state AD generation as the following optimization problem:
\begin{argmaxi!}|s|
{\mathbf{x}, \mathbf{y}, \mathbf{d}}{
\sum_{i} \sum_{k=0}^{K} \sigma_{i}\, l_{ik}\, y_{ik}
\label{opt:sco}}{}{}
\addConstraint{\sum_{k=0}^{K}y_{ik}=}{ \ x_i \quad}{\forall i \label{con:com}}
\addConstraint{[d_i, d_i + L_i] \subseteq}{ \ \mathcal{G} \quad}{\forall i : x_i = 1\label{con:gap}}
\addConstraint{\left|  d_i + \tfrac{L_i}{2} - m_i \right| \leq}{ \ \Delta_{\text{max}} \quad}{\forall i : x_i = 1\label{con:occ}}
\addConstraint{d_i+L_i\leq}{ \ d_u \quad}{\forall\, i<u:x_i=x_u=1\label{con:ove}}
\end{argmaxi!}

The objective \labelcref{opt:sco} maximizes  salience  weighted by the narration time, so   each second of a gap accrues the salience of the eventuality it describes. Within an eventuality this favours the longest variant that fits; without the weighting the solver would be indifferent between a full description and its shortest compression;  
 across eventualities, a dense short description displaces a longer, less salient one 
   whenever the time it frees can be filled at least as densely.
The assumption that such content is available is met for AD: movies depict far more eventualities than can be narrated.

The objective is subject to four constraints.
The first equation~\labelcref{con:com} concerns compression: a described eventuality uses exactly one variant, and an omitted one uses none.
The remaining three concern temporal placement.
They require each description to fall within a permissible narration interval \labelcref{con:gap}, to remain temporally close to the eventuality it describes \labelcref{con:occ}, and not to overlap the next description, which also preserves the order of the scene description \labelcref{con:ove}.
Constraints conditioned on \(x_i = 1\) are indicator constraints, and equation~\labelcref{con:gap} is a disjunction over intervals; Section~\ref{sec:setup} describes how both are linearized.
The formulation is agnostic as to how its inputs are obtained; Section~\ref{sec:setup} describes our choices for scene description, occurrence spans, compression, and salience scoring.

\begin{figure}[t]
\centering
\resizebox{\columnwidth}{!}{%
\begin{tikzpicture}[x=0.95cm, y=1cm, font=\scriptsize, >=Latex,
  ev/.style={draw=reframed-gold, fill=reframed-yellow!25, rounded corners=1.5pt},
  dia/.style={draw=reframed-silver, fill=reframed-gray!70},
  gap/.style={draw=reframed-green!60, fill=reframed-green!8},
  ad/.style={draw=reframed-green!70!black, fill=reframed-green!12, rounded corners=1.5pt},
  lbl/.style={font=\scriptsize},
  tny/.style={font=\tiny},
]

\node[anchor=east, lbl] at (-0.25,3.05) {\textbf{Video}};
\node[anchor=east, lbl] at (-0.25,1.75) {\textbf{Soundtrack}};
\node[anchor=east, lbl, align=center] at (-0.25,0.35) {\textbf{Audio}\\[-2pt] \textbf{Description}};

\begin{scope}[on background layer]
  \fill[reframed-gold!12] (8,2.55) rectangle (12,3.30);
\end{scope}
\draw[reframed-gold!70, decorate, decoration={brace, amplitude=3pt}] (12,3.34) -- (8,3.34);
\node[tny, text=reframed-gold!80!black, anchor=south] at (10,3.46)
  {$|d_4+\tfrac{L_4}{2}-m_4|\le\Delta_{\max}$};

\draw[ev] (1,2.85) rectangle (4,3.25);   \node[lbl] at (2.5,3.05) {$e_1$ parks};
\draw[ev] (4,2.85) rectangle (6,3.25);   \node[lbl] at (5,3.05) {$e_2$ helmet};
\draw[ev] (6,2.85) rectangle (8,3.25);   \node[lbl] at (7,3.05) {$e_3$ package};
\draw[ev] (8,2.85) rectangle (12,3.25);  \node[lbl] at (10,3.05) {$e_4$ sees Mikael};
\foreach \m/\i in {2.5/1, 5/2, 7/3, 10/4}{
  \draw[reframed-gold, thick] (\m,2.83) -- (\m,2.63);
  \node[tny, text=reframed-gold!80!black, anchor=north, inner sep=1pt] at (\m,2.65) {$m_{\i}$};
}

\draw[dia] (0,1.55) rectangle (1.5,1.95);   \node[lbl] at (0.75,1.75) {dialogue};
\draw[gap] (1.5,1.55) rectangle (11,1.95);
\node[lbl, text=reframed-green!60!black] at (6.25,1.75)
  {$\mathcal{G}$: 9.5\,s of narration time between dialogue};
\draw[dia] (11,1.55) rectangle (13,1.95);   \node[lbl] at (12,1.75) {dialogue};

\draw[ad] (1.5,0.15) rectangle (4.9,0.55);
\node[lbl] at (3.2,0.35) {``She parks on a corner\dots''};
\node[tny, anchor=north] at (3.2,0.13) {$x_1{=}1$, $L_1{=}3.3$\,s};
\draw[ad] (4.9,0.15) rectangle (7.8,0.55);
\node[lbl] at (6.35,0.35) {``takes a package\dots''};
\node[tny, anchor=north] at (6.35,0.13) {$x_3{=}1$, $L_3{=}3.0$\,s};
\draw[ad] (7.8,0.15) rectangle (11,0.55);
\node[lbl] at (9.4,0.35) {``stops when she sees\dots''};
\node[tny, anchor=north] at (9.4,0.13) {$x_4{=}1$, $L{_4=}2.7$\,s};

\draw[reframed-gold!80, dotted, thick] (9.4,0.6) -- (9.4,2.55);
\node[tny, text=reframed-gold!80!black, anchor=west, inner sep=1.5pt] at (9.45,1.2) {delivery midpoint};

\draw[->, reframed-red!70] (5,3.62) -- (5,3.29);
\node[lbl, text=reframed-red, anchor=south, align=center] at (5,3.64)
  {$e_2$ dropped: $x_2{=}0$\\[-2pt] \tiny lowest salience, no room in $\mathcal{G}$};

\node[draw=reframed-silver, fill=black!2, rounded corners=2pt, anchor=north west,
      align=left, inner sep=4pt, lbl] at (current bounding box.west |- 0,-1.05)
  {The candidate wordings for $e_4$ are
   $\mathcal{C}_4=\{c_{4,0},\dots,c_{4,3}\}$: the full description element $w_4 = c_{4,0}$ extracted from a\\scene description and three progressively shorter compressions of it. Exactly one is selected, by $y_{4k}$.\\[3pt]
   \begin{tabular}{@{}lllll@{}}
   $c_{4,0}$ & ``but stops dead when she sees Mikael leave his apartment with Erika''
     & $l_{4,0}=3.6$\,s &   & \textcolor{reframed-red}{too long for $\mathcal{G}$}\\
   $c_{4,1}$ & ``but stops when she sees Mikael leave with Erika''
     & $l_{4,1} = 2.7$\,s &  & \textcolor{reframed-green}{selected: $y_{4,1}{=}1$, so $L_4=l_{4,1}$}\\
   $c_{4,2}$ & ``but stops when she sees Mikael and Erika''
     & $l_{4,2}= 2.4$\,s &   & fits, but loses \emph{leaving}\\
   $c_{4,3}$ & ``she stops dead''
     & $l_{4,3}=0.9$\,s &   & fits, but loses who she sees\\
   \end{tabular}   };

\draw[reframed-silver] (0,-0.55) -- (13,-0.55);
\foreach \t/\lab in {0/44, 2/46, 4/48, 6/50, 8/52, 10/54, 12/56}{
  \draw[reframed-silver] (\t,-0.55) -- (\t,-0.68);
  \node[tny, anchor=north, text=black!60] at (\t,-0.68) {\lab\,s};
}
\end{tikzpicture}%
}
\caption{Example of the optimization, on the final scene of Figure~\ref{fig:dragon-tattoo}. Four eventualities occur on screen (top), each with an occurrence midpoint $m_i$, while the soundtrack leaves one gap $\mathcal{G}$ between lines of dialogue (middle). The solution (bottom) makes all three decisions jointly: it sets $x_2{=}0$, dropping the least salient eventuality because the others cannot otherwise fit; it chooses delivery times $d_i$ that tile $\mathcal{G}$ in order and without overlap; and it selects the compression $c_{4,1}$ for $e_4$: the full description $c_{4,0}$ would exceed the time left, and because the objective weights salience by narration time, the longest variant that fits scores highest. The shaded band shows the $\Delta_{\max}$ window that keeps a description near the event it describes. Durations and gap boundaries are illustrative.}
\label{fig:optimization}
\end{figure}
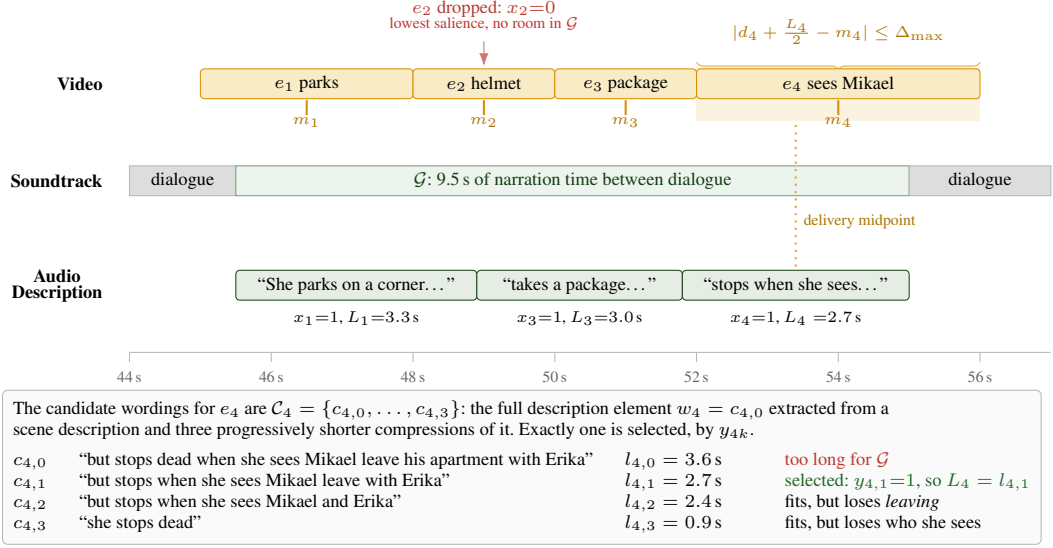

\textbf{AD Generation Example.} Figure~\ref{fig:optimization} illustrates the optimization on the final scene of Figure~\ref{fig:dragon-tattoo}. Four eventualities occur on screen, while dialogue leaves a single gap of 9.5 seconds, which is too short to narrate all of them in full.  The solver resolves this conflict through all three decisions at once.  It drops the least salient eventuality, Lisbeth removing her helmet ($x_2=0$); it packs the remaining descriptions within the gap, in order and each within $\Delta_{\max}$ of its occurrence midpoint; and, since the full description of $e_4$ would overrun the gap, it selects a compression ($y_{4,1}=1$), the longest variant that still fits. Appendix~\ref{app:worked-example} walks through input preparation and the solution for a longer scene.

\section{Experimental Setup}\label{sec:setup}

This section describes our data and evaluation protocol.  %
We then describe how the MILP system is implemented, including its inputs, salience scoring alternatives, and the systems we compare against.

\textbf{Data.} We use REFRAMED \citep{sterner-2026-reframed}, a dataset of 2,023 video excerpts (average length 144s) spanning 3,302 scenes from 206 movies.  Each excerpt comes with two professional AD versions (American and British), professional dialogue subtitles, and, for 85 movies, screenplays aligned at the scene level.
We report main results on the REFRAMED challenge set: ten full-length movies, manually annotated with scene boundaries and with two to three professionally transcribed AD versions each.
The MILP is solved per scene; there are a total of 1{,}221 scenes, mean length is~57.4s.
Ablations use a separate validation set of eight movies from the REFRAMED training and validation splits, chosen as those whose screenplays best match the final post-production movie.\footnote{For our method for scoring the degree of match and an analysis of the correlation of this measure against a QA-based evaluation of the screenplay scene descriptions, see Appendix~\ref{sub:screenplays}.}  Screenplay scene descriptions are written to guide production, so they emphasize narrative intent rather than details fixed at the time of shooting, which makes it interesting to compare MILP inputs derived from screenplays against LLM-generated scene descriptions.

\textbf{Evaluation.} We follow the REFRAMED evaluation protocol, which compares generations against multiple references with three groups of metrics (details in Appendix~\ref{app:metrics}). \emph{Dialogue-gap} metrics compute CIDEr and METEOR between generated and reference descriptions assigned to each gap in the dialogue subtitles.  \emph{Alignment} metrics use SODA \citep{fujita-2020-soda} to align generated and reference descriptions monotonically: SODA-M is the METEOR score of aligned pairs, and SODA-T the proportion of references aligned within $\tau=10$ seconds. \emph{QA-based} metrics measure narrative usefulness: QEval is the proportion of reference-based questions generated AD answers correctly; QEval-T requires the answer comes from a description within $\tau$ seconds of the correct time.  Statistical significance is assessed with two-tailed Monte Carlo permutation tests ($R=10{,}000$, $\alpha=0.05$).

Prompted LLMs can be rewarded for descriptions that in practice cannot be narrated.
We therefore also report \textbf{realistic} variants (the indented rows in Table~\ref{tab:challenge-results}), which remove descriptions that would need to be narrated faster than 300\,WPM,\footnote{The LLM prompts request 180\,WPM, the rate in American AD guidelines \citep{adc_2009_guidelines}; over 99\% of REFRAMED reference
descriptions are delivered below 300\,WPM.} fall outside a dialogue gap (with a one-second collar), or begin before the preceding description ends.
The filter applies only to the prompted LLMs.
Our MILP satisfies these conditions by construction: constraint \eqref{con:gap} places every description inside a dialogue gap, constraint \eqref{con:ove} prevents descriptions from overlapping, and narration durations are computed at 200\,WPM, so no description is narrated faster than it can be spoken.

\paragraph{MILP system.}

We build two versions of the system, in which all MILP inputs are produced by either Qwen 3.5 27B \citep{qwen3.5} or Gemini 3.1 Flash-Lite \citep{gemini3.1flashlite}. Video is sampled at one frame per second, without audio. The Qwen-based pipeline runs locally with vLLM on two H200 GPUs; Gemini is accessed through the official paid API.

The LLM generates a prose scene description from the video (prompt in Appendix~\ref{sec:screenplays:generation}). Because video input alone does not support correct character naming, we augment it with two sources of automatically computed character information: character names with face crops (our IMDb-portrait comparison method achieves $F_1=73.7$; Appendix~\ref{sub:character_face_detection}), and dialogue timestamps labeled with the speaking character (our speaker diarization method achieves $F_1=77.9$; Appendix~\ref{sec:identification}).
The scene description is segmented into sentences \citep{frohmann-2024-segment} and then into description elements using REFRAMED's learned segmentation model (character-level $F_1=66.0$ vs.\ $38.4$ for a comma-splitting baseline).
A second prompt (Appendix~\ref{sec:llm-processing}) takes the video and the full sequence of description elements, and returns for each element its occurrence span and five compressed variants, at 0.9, 0.8, 0.7, 0.6 and 0.5 times the source length.  The same prompt also produces our default salience scores, assigning each element a score in $[0,1]$; salience represents the extent to which the element provides information necessary for following the story, relative to the other elements of the scene.  Each variant is assigned a narration duration at a fixed speaking rate of 200 words per minute, computed from the generated words. Permissible narration times $\mathcal{G}$ are the gaps of at least one second between REFRAMED dialogue subtitles.

Because salience drives content selection, we compare our LLM scores against four alternatives and two oracles, leaving the rest of the pipeline unchanged.  \textbf{Random} draws each score uniformly from $[0,1]$.  The other three follow \citet{sterner-2026-contrastive}, who define salience as the similarity between an element and a representation of the narrative as a whole.  \textbf{Embedding video} and \textbf{Embedding text} take the cosine similarity between the element embedding and an embedding of the full video and of the full scene description, respectively, using Qwen 3 VL Embedding (8B) over the Qwen descriptions;  Video embeddings are native, with the video-trained model encoding all frames jointly.
\textbf{BM25} scores the lexical overlap between the element and the scene description.  The two oracles indicate how much headroom better salience estimation can contribute, by scoring each element against the reference AD: \textbf{BM25 oracle} and \textbf{Embedding oracle} use BM25 and embedding similarity, respectively. 
Salience scores are not normalized; empirically, we find all predictions are non-negative.

The MILP is solved with Gurobi 13.0.3 \citep{gurobi} on a node with 24 CPUs. Gap containment \eqref{con:gap} is linearized with binary gap-assignment variables, and the constraints conditioned on $x_i=1$ are encoded as Gurobi indicator constraints.
We set $\Delta_{\max}=10$ seconds, following the analysis of manually timecoded AD in \citet{sterner-2026-reframed}, which shows that professional describers narrate an element within ten seconds of its occurrence (in 99.4\% of cases).  %
Solving stops when Gurobi's default optimality criteria are met or after ten minutes, whichever comes first.
Of the scenes solved optimally, mean and median solve times are 2.75s and 0.12s, respectively.
The solving for 11 and 15 scenes was stopped due to the ten-minute timeout, for Qwen and Gemini inputs respectively; these have a mean optimality gap of 18.1\% and 19.2\%.
The conditions that lead to timeout are a complex combination of the objective and constraints: for instance, one scene with Qwen inputs reaches the timeout despite only being 32 seconds long, a result of the 33 candidate eventualities needing to be scheduled into only two dialogue gaps of total length 14.8s.

\textbf{Comparison systems.} We compare against the published generations and results of \citet{sterner-2026-reframed}, reporting their best variant under the evaluation protocol: \textbf{Qwen 3.5} and \textbf{Gemini 3.1}, the same models used to produce the MILP inputs, here prompted with the video to generate timecoded descriptions directly.
Their model is given the video together with the dialogue subtitles and the list of gaps of at least one second, and is not told which gaps should contain a description.
The lower bound is a greedy \textbf{Random} baseline that fills each dialogue gap with randomly sampled training-set description elements until no further element fits.  The upper bound is an \textbf{Expert} AD transcript, evaluated as a system output.  \textbf{ShotbyShot} \citep{xie-2025-shot_by_shot} and \textbf{DistinctAD} \citep{fang-2025-distinct_ad} are specialist AD systems that generate one description sentence from one clip and its context.

Finally, a controlled comparison replaces the solver with an LLM. The \textbf{Qwen scheduler} and \textbf{Gemini scheduler} receive the same information as the MILP (description elements, occurrence spans, salience scores, 200\,WPM narration durations, and dialogue gaps), together with the text of the elements and their compressions, which the MILP does not use. They produce the same outputs as the MILP, a selected variant and a delivery start time for each element, relying on the LLM's parametric knowledge of AD rather than the symbolic constraints given to the MILP. We run Qwen 3.5 with reasoning enabled and Gemini 3.1 with \texttt{thinking=high}, as in all other LLM runs.

\section{Results and Discussion}\label{sec:results}

\begin{figure}[t]
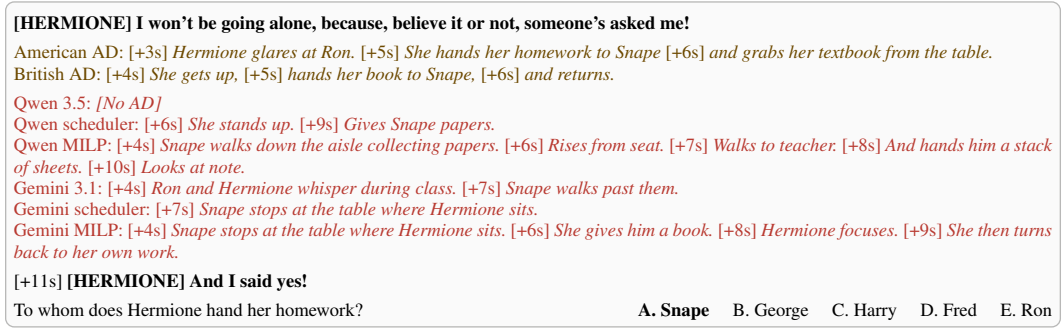


\begin{tcolorbox}[
  colback=reframed-gray!10,
  colframe=reframed-silver,
  boxrule=0.5pt,
  arc=3pt,
  left=0pt, right=0pt, top=8pt, bottom=8pt,
  width=\columnwidth,
  fontupper=\scriptsize
]
\vspace{-.2cm}
\textbf{[HERMIONE] I won't be going alone, because, believe it or not, someone's asked me!}

\smallskip

{\color{reframed-gold!50!black}American AD: [+3s] \textit{Hermione glares at Ron.} [+5s] \textit{She hands her homework to Snape} [+6s] \textit{and grabs her textbook from the table.}\par}
{\color{reframed-gold!50!black}British AD: [+4s] \textit{She gets up,} [+5s] \textit{hands her book to Snape,} [+6s] \textit{and returns.}\par}

\smallskip

{\color{reframed-red}Qwen 3.5: \textit{[No AD]}\par}
{\color{reframed-red}Qwen scheduler: [+6s] \textit{She stands up.} [+9s] \textit{Gives Snape papers.}\par}
{\color{reframed-red}Qwen MILP: [+4s] \textit{Snape walks down the aisle collecting papers.} [+6s] \textit{Rises from seat.} [+7s] \textit{Walks to teacher.} [+8s] \textit{And hands him a stack of sheets.} [+10s] \textit{Looks at note.}\par}
{\color{reframed-red}Gemini 3.1: [+4s] \textit{Ron and Hermione whisper during class.} [+7s] \textit{Snape walks past them.}\par}
{\color{reframed-red}Gemini scheduler: [+7s] \textit{Snape stops at the table where Hermione sits.}\par}
{\color{reframed-red}Gemini MILP: [+4s] \textit{Snape stops at the table where Hermione sits.} [+6s] \textit{She gives him a book.} [+8s] \textit{Hermione focuses.} [+9s] \textit{She then turns back to her own work.}\par}

\smallskip

[+11s] \textbf{[HERMIONE] And I said yes!}

\smallskip

To whom does Hermione hand her homework? \hfill \textbf{A. Snape} \quad B. George \quad C. Harry \quad D. Fred \quad E. Ron\par

\vspace{-.3cm}
\end{tcolorbox}
\vspace{-8pt}
\caption{\textcolor{reframed-gold!50!black}{References} and \textcolor{reframed-red}{generated ADs} for dialogue gap in \textit{Harry Potter and the Goblet of Fire} (2005). Dialogue in \textbf{bold}, AD in \emph{italics}. Multiple-choice question  is from the QA-based evaluation.}
\label{ex:harry}
\vspace{-15pt}
\end{figure}

\textbf{Does the MILP reach state-of-the-art automatic AD?}
Table~\ref{tab:challenge-results} reports results on the REFRAMED
\begin{wraptable}[23]{r}{0.5\textwidth}
\vspace{-8pt}
\centering
\caption{System performance on REFRAMED challenge set with  six evaluation metrics. Indented rows apply the realistic filter. (Dialogue) gap columns are CIDEr and METEOR; SODA columns are SODA-M and SODA-T; QEval columns are QEval (Acc) and QEval-T (T).}
\label{tab:challenge-results}
\begin{small}
\setlength{\tabcolsep}{3pt}
\vspace{-.2cm}
\begin{tabular}{@{}l *{6}{S[table-format=2.1, table-column-width=0.56cm]}@{}}
\toprule
& \multicolumn{2}{c}{\bf Gaps} & \multicolumn{2}{c}{\bf SODA} & \multicolumn{2}{c}{\bf QEval} \\
\cmidrule(lr){2-3} \cmidrule(lr){4-5} \cmidrule(lr){6-7}
& \multicolumn{1}{c}{\bf C} & \multicolumn{1}{c}{\bf M} & \multicolumn{1}{c}{\bf M} & \multicolumn{1}{c}{\bf T} & \multicolumn{1}{c}{\bf Acc} & \multicolumn{1}{c}{\bf T} \\
\midrule
Expert & 51.4  & 20.1 & 16.3  & 81.0 & 69.6 & 61.2  \\ \midrule
Random & 1.3 & 4.4 & 4.0 & 27.1 & 34.1 & 2.3 \\
DistinctAD & 15.6 & 7.1 & 8.3 & 49.0 & 35.7 & 8.8 \\ 
ShotbyShot & 16.3 & 7.0 & 8.0 & 53.0 & 39.9 & 17.0 \\ 
Qwen 3.5 & 13.2 & 8.1 & 7.4 & 38.4 & 43.9 & 17.3 \\
\quad realistic & 11.3 & 6.5 & 7.1 & 31.2 & 42.6 & 14.8 \\ 
Gemini 3.1 & 19.0 & 8.2 & 7.9 & 36.0 & 42.3 & 16.4 \\
\quad realistic & 16.9 & 7.3 & 7.7 & 32.0 & 41.9 & 15.1 \\  \midrule
Qwen scheduler & 14.0 & 7.1 & 7.2 & 47.6 & 43.3 & 21.8 \\
Qwen MILP & 11.4 & 7.9 & 7.7 & 53.6 & 43.6 & 22.1 \\ \midrule
Gemini scheduler & 14.5 & 6.6 & 6.9 & 45.8 & 41.6 & 19.9 \\
Gemini MILP & 13.6 & 8.4 & 8.0 & 55.7 & 45.9 & 25.5 \\
\bottomrule
\end{tabular}
\end{small}
\end{wraptable}
challenge set. Among automatic systems, Gemini MILP performs best on both QA-based metrics (QEval=45.9 and QEval-T=25.5; $p<0.01$ against every other automatic system). It also 
achieves the best SODA-T score except for ShotbyShot, with which the difference is not significant. Qwen MILP shows the same pattern: its largest gains over realistic Qwen are on SODA-T (53.6 vs.\ 31.2) and QEval-T (22.1 vs.\ 14.8; both $p<0.01$). DistinctAD and ShotbyShot are advantaged on SODA-T (49.0 and 53.0) because they inherit the reference AD's placement and content selection, but this does not translate into comparable narrative performance (QEval-T=8.8 and 17.0). MILP systems do not improve CIDEr (e.g., Gemini MILP 13.6 vs.\ Gemini 3.1 19.0), which is not unexpected: CIDEr rewards n-gram overlap with the reference wording, which favours the fluent full sentences the prompted LLMs produce, whereas the MILP's compression and selection yield terser, sometimes fragmentary descriptions (see Figure~\ref{ex:harry}) that convey the same content in different words. The gain on QA-based metrics alongside the CIDEr drop reflects this: the MILP trades surface overlap with a single reference for narrative usefulness. Under all evaluation metrics, the expert human upper bound significantly outperforms all other systems (all $p<0.01$), and the random baseline is outperformed by all other systems (all $p<0.01$).

\textbf{Does explicit optimization outperform an LLM scheduler?}
Under the controlled comparison between the MILP and LLM schedulers, the MILP improves results for both Gemini and Qwen on all but the lexical metrics.
The MILP is significantly better under every metric except CIDEr, on which it is indistinguishable from the Gemini scheduler and worse than the Qwen scheduler ($p<0.01$); all other differences have $p<0.05$. 
The gains are largest on temporal placement, and larger for Gemini than for Qwen: Gemini MILP improves on the Gemini scheduler under both temporal metrics (SODA-T 55.7 vs.\ 45.8; QEval-T 25.5 vs.\ 19.9), whereas Qwen differences are concentrated in SODA-T (53.6 vs.\ 47.6), with only numerically small gains on QEval-T (22.1 vs.\ 21.8). 
These results show that explicit optimization improves coordination of the \emph{what}, \emph{when}, and \emph{how} decisions, while the scheduler's strong performance suggests that the staged representation is helpful.

On average the MILP selects 45.2\% of candidate eventualities for Qwen and 41.2\% for Gemini, and it selects fewer when narration time is scarcer. For Gemini, the share selected rises steadily with available narration time, from 21.0\% 
\begin{wraptable}[15]{r}{0.45\textwidth}
\vspace{-8pt}
\centering
\caption{Performance stratified by MILP solve time, an empirical proxy for \emph{AD difficulty}. As scenes  become harder to schedule (longer runtimes),  scores drop, while the MILP's advantage ($\Delta$) over the LLM scheduler widens.}
\label{tab:runtime}
\vspace{-.2cm}
\begin{small}
\setlength{\tabcolsep}{3pt}
\begin{tabular}{@{}l *{6}{S[table-format=2.1, table-column-width=0.63cm]}@{}}
\toprule
& \multicolumn{3}{c}{\bf SODA-T} & \multicolumn{3}{c}{\bf QEval} \\
\cmidrule(lr){2-4} \cmidrule(lr){5-7}
\makecell[l]{Solve\\ time}& LLM & MILP & $\Delta$ & LLM & MILP & $\Delta$ \\
\midrule
     $<$ 0.1s & 57.5 & 62.4 & +4.9 & 48.3 & 51.4 & +3.1 \\
     0.1--1s    & 46.9 & 57.0 & +10.1 & 42.2 & 45.8 & +3.6 \\
     1--10s     & 45.0 & 55.6 & +10.6 & 43.7 & 48.3 & +4.6 \\
     10--60s    & 37.8 & 50.1 & +12.3 & 33.9 & 40.8 & +6.9 \\
\bottomrule
\end{tabular}
\end{small}
\end{wraptable}
when under 5s is free, through 32.7\%, 48.7\%, 67.2\% and 80.6\%, to 91.8\% when more than 80s is available.
 Among the selected eventualities, 49.4\% keep their full variant and the rest are compressed, with the compressions skewed toward lighter rates (0.9: 18.6\%, 0.8: 9.7\%, 0.7: 5.0\%, 0.6: 6.2\%, 0.5: 11.2\%). This follows from the length-weighted objective, which favours the full variant when there is space but is forced toward heavier compression, and hence the tail at 0.5, when gaps are tight.
Table~\ref{tab:runtime} turns to how SODA-T and QEval vary with solve time, for both the MILP and the LLM scheduler. The MILP's advantage ($\Delta$) grows as solve time increases, showing the solver's strength precisely when selection, placement, and compression must be tightly coordinated.

\textbf{Do LLMs game the AD metrics?} 
The realistic variants (Table~\ref{tab:challenge-results}) are significantly worse than raw LLMs across all metrics (all $p<0.01$). Dialogue overlaps are rare (1.6\% and 1.2\% of generations for Qwen 3.5 and Gemini 3.1) and no two descriptions overlap each other; the dominant failure is excessive speaking rate, with 28.9\% and 16.7\% of generations exceeding the 300\,WPM ceiling, and a third and a tenth of those respectively exceeding even 500\,WPM. These LLMs thus lack the temporal and linguistic control AD requires, and the realistic filter prevents them from exploiting the metrics with descriptions that could not be narrated.  The MILP satisfies the filter by construction, so it never alters results; the LLM scheduler is likewise reported after filtering for fairness, with near-identical numbers (dialogue-gap scores are unchanged and all other differences are $\leq$ 0.4 points).

Consider Figure~\ref{ex:harry}, which shows a seven second dialogue gap half way into a \textit{Harry Potter} movie.
Snape is walking the aisles of a classroom, as students work.
Hermione reveals that someone has asked her to the upcoming Christmas ball, then gets up, hands her homework to Snape, and returns to say that she has accepted.
The prompted Qwen 3.5 does not generate anything for this dialogue gap, while Gemini describes the characters whispering and Snape walking past (which occurs much earlier).
The Qwen scheduler is better, but using only six words is extremely terse, while the Gemini scheduler selects only a single 2 second description about Snape.
Meanwhile, both MILPs describe Hermione giving Snape her homework, which is what the relevant evaluation QA pair tests.
Note that as a result of the selection decisions in the MILP, the generations are not perfectly fluent, and the Qwen MILP uses an incomplete sentence (\textit{Rises from seat}).

\textbf{Which components contribute to performance?} Table~\ref{tab:ablation-results} ablates Qwen MILP on the validation set. Removing the grounding constraint (row~2) primarily hurts temporal performance (SODA-T
  \begin{wraptable}[24]{r}{0.55\textwidth}
\vspace{-8pt}
\centering
\caption{Qwen performance on  REFRAMED post-production screenplay set using \textbf{C}IDEr and  \textbf{M}ETEOR,  SODA-\textbf{M}, SODA-\textbf{T}, QEval (\textbf{Acc}), and QEval-\textbf{T}.}
\label{tab:ablation-results}
\begin{small}
\setlength{\tabcolsep}{3pt}
\vspace{-.1cm}
\begin{tabular}{@{}l *{6}{S[table-format=2.1, table-column-width=0.6cm]}@{}}
\toprule
& \multicolumn{2}{c}{\bf Gaps} & \multicolumn{2}{c}{\bf SODA} & \multicolumn{2}{c}{\bf QEval} \\
\cmidrule(lr){2-3} \cmidrule(lr){4-5} \cmidrule(lr){6-7}
& \multicolumn{1}{c}{\bf C} & \multicolumn{1}{c}{\bf M} & \multicolumn{1}{c}{\bf M} & \multicolumn{1}{c}{\bf T} & \multicolumn{1}{c}{\bf Acc} & \multicolumn{1}{c}{\bf T} \\
\midrule
(1) Qwen MILP & 12.9 & 7.7 & 7.3 & 54.9 & 48.4 & 28.2 \\ \midrule
\multicolumn{7}{c}{\bf MILP decisions and constraints} \\ \midrule
(2) w/o grounding  & 11.0 & 7.0 & 7.4 & 47.8 & 44.7 & 16.0 \\
(3) w/o compression  & 9.2 & 6.2 & 7.3 & 43.7 & 43.4 & 22.7 \\
(4) + w/o both  & 6.6 & 6.0 & 7.6 & 41.0 & 39.7 & 12.9 \\ \midrule
\multicolumn{7}{c}{\bf Salience scoring} \\ \midrule
(5) Random & 12.1 & 7.2 & 7.1 & 53.1 & 43.6 & 20.0  \\
(6) Embed. video & 11.2 & 6.9 & 6.9 & 51.3 & 41.6 & 20.4 \\
(7) Embed. text  & 12.3 & 7.3 & 7.3 & 50.7 & 43.8 & 22.5\\
(8) BM25 & 13.5 & 7.6 & 7.3 & 54.2 & 46.0 & 22.2 \\ \midrule
(9) BM25 oracle  & 16.3 & 8.6 & 8.0 & 51.4 & 46.6 & 23.8 \\
(10) Embed. oracle & 16.5 & 8.8 & 7.8 & 54.8 & 49.9 & 24.2 \\ \midrule
\multicolumn{7}{c}{\bf Scene descriptions} \\ \midrule
(11) Screenplay & 13.5 & 7.3 & 7.4 & 47.7 & 53.1 & 31.1 \\
\bottomrule
\end{tabular}%
\end{small}
\end{wraptable}
54.9 $\rightarrow$ 47.8; QEval-T 28.2 $\rightarrow$ 16.0), while removing compressed variants (row~3) reduces content-based metrics (CIDEr 12.9 $\rightarrow$ 9.2; QEval 48.4 $\rightarrow$ 43.4), because fewer descriptions can fit the available gaps. 
 Removing both gives the worst overall performance (row~4).
 Thus, grounding determines where descriptions can be placed; compression increases how much useful content can be retained.

Rows~(5)--(10) vary salience scoring, leaving the remaining pipeline fixed. Non-oracle methods differ little on lexical and temporal metrics, but separate more clearly on QA: LLM salience reaches QEval=48.4 and QEval-T=28.2, compared with 43.6 and 20.0 for random salience. Better salience therefore primarily improves \emph{what} is selected rather than \emph{when} it is delivered. The oracle scores support this: they improve CIDEr (16.3--16.5 vs.\ 12.9) and not SODA-T or QEval-T.

Finally, replacing the LLM scene description with screenplay descriptions (row~11) gives the best QA performance (QEval=53.1; QEval-T=31.1), suggesting screenplays better capture narrative intent. Temporal alignment falls (SODA-T 54.9 $\rightarrow$ 47.7), plausibly because pre-production screenplay events are harder to ground in the final video. Appendix~\ref{sec:social-network} provides qualitative examples.

\section{Conclusion}\label{sec:conclusion}

We formalized realistic AD generation as a constrained optimization problem over what to describe, when to deliver it, and how to formulate it.
We instantiated this view in a hybrid system in which LLMs propose, ground, score, and compress visual events while a mixed-integer linear program makes joint decisions across a scene. On REFRAMED, the resulting system achieves the strongest QA-based performance among automatic systems and improves over an LLM scheduler given the same structured inputs. The gains are concentrated in temporal placement and narrative usefulness rather than n-gram overlap, reflecting the aspects of AD that the formulation explicitly controls.

Our analysis ablations clarify where these gains come from.
Temporal grounding and compression are critical for placing descriptions within the limited narration time, while salience primarily determines which parts of the story are retained. Oracle salience scores reveal further headroom in content selection, although they do not consistently improve temporal placement or QA-based measures;  screenplay-based scene descriptions,  by contrast,  yield the largest gains on QA-based measures, suggesting that representing narrative importance remains a major bottleneck. At the same time, a substantial gap to professional describers remains in both content selection and temporal realization. Closing it will require better models of narrative salience and visual grounding; importantly, these improvements can be incorporated as better inputs to the same global optimization framework.

\subsection*{AI use statement}

In this work, we used generative AI tools for assistance with coding implementation and for writing feedback.
We have not used generative AI tools for any of the following: generating synthetic data (all reference AD used is human-authored), help with the conceptual development of our framework, formulation of mathematical claims, assistance in the writing of proofs, proposals of hypotheses, design of our research methodology or experiments, translation, data cleaning, or results interpretation.
We have reviewed all AI-assisted work manually.
We take responsibility for the final content of this work, including text, claims or artifacts produced with the aid of generative AI.

\bibliography{iclr2027_conference}

\begin{thebibliography}{49}
\providecommand{\natexlab}[1]{#1}
\providecommand{\url}[1]{\texttt{#1}}
\expandafter\ifx\csname urlstyle\endcsname\relax
  \providecommand{\doi}[1]{doi: #1}\else
  \providecommand{\doi}{doi: \begingroup \urlstyle{rm}\Url}\fi

\bibitem[Alayrac et~al.(2022)Alayrac, Donahue, Luc, Miech, Barr, Hasson, Lenc, Mensch, Millican, Reynolds, Ring, Rutherford, Cabi, Han, Gong, Samangooei, Monteiro, Menick, Borgeaud, Brock, Nematzadeh, Sharifzadeh, Bi\'{n}kowski, Barreira, Vinyals, Zisserman, and Simonyan]{alayrac-2022-flamingo}
Jean-Baptiste Alayrac, Jeff Donahue, Pauline Luc, Antoine Miech, Iain Barr, Yana Hasson, Karel Lenc, Arthur Mensch, Katherine Millican, Malcolm Reynolds, Roman Ring, Eliza Rutherford, Serkan Cabi, Tengda Han, Zhitao Gong, Sina Samangooei, Marianne Monteiro, Jacob~L Menick, Sebastian Borgeaud, Andy Brock, Aida Nematzadeh, Sahand Sharifzadeh, Miko\l~aj Bi\'{n}kowski, Ricardo Barreira, Oriol Vinyals, Andrew Zisserman, and Kar\'{e}n Simonyan.
\newblock Flamingo: a visual language model for few-shot learning.
\newblock In S.~Koyejo, S.~Mohamed, A.~Agarwal, D.~Belgrave, K.~Cho, and A.~Oh (eds.), \emph{Advances in Neural Information Processing Systems}, volume~35, pp.\  23716--23736. Curran Associates, Inc., 2022.

\bibitem[{Audio Description Coalition}(2009)]{adc_2009_guidelines}
{Audio Description Coalition}.
\newblock \emph{Standards for Audio Description and Code of Professional Conduct for Describers based on the training and experience of audio describers and trainers from across the United States}, 3rd edition, 2009.
\newblock URL \url{https://audiodescriptionsolutions.com/wp-content/uploads/2020/04/adc_standards_090615.pdf}.

\bibitem[Bach(1986)]{bach-1986-eventuality}
Emmon Bach.
\newblock The algebra of events.
\newblock \emph{Linguistics and Philosophy}, 9\penalty0 (1):\penalty0 5--16, 1986.
\newblock ISSN 01650157, 15730549.
\newblock URL \url{http://www.jstor.org/stable/25001229}.

\bibitem[Berg-Kirkpatrick et~al.(2011)Berg-Kirkpatrick, Gillick, and Klein]{berg-2011-jointly}
Taylor Berg-Kirkpatrick, Dan Gillick, and Dan Klein.
\newblock Jointly learning to extract and compress.
\newblock In Dekang Lin, Yuji Matsumoto, and Rada Mihalcea (eds.), \emph{Proceedings of the 49th Annual Meeting of the Association for Computational Linguistics: Human Language Technologies}, pp.\  481--490, Portland, Oregon, USA, June 2011. Association for Computational Linguistics.
\newblock URL \url{https://aclanthology.org/P11-1049/}.

\bibitem[Chai et~al.(2025)Chai, Song, Du, Meng, Madhavan, Bar-Tal, Hwang, Xie, and Manning]{chai-2025-auroracap}
Wenhao Chai, Enxin Song, Yilun Du, Chenlin Meng, Vashisht Madhavan, Omer Bar-Tal, Jenq-Neng Hwang, Saining Xie, and Christopher~D Manning.
\newblock {AuroraCap}: Efficient, performant video detailed captioning and a new benchmark.
\newblock In \emph{The Thirteenth International Conference on Learning Representations (ICLR)}, 2025.
\newblock URL \url{https://openreview.net/forum?id=tTDUrseRRU}.

\bibitem[Chatman(1980)]{chatman-1980-novels}
Seymour Chatman.
\newblock What novels can do that films can't (and vice versa).
\newblock \emph{Critical Inquiry}, 7:\penalty0 121--140, 1980.
\newblock URL \url{https://doi.org/10.1086/448091}.

\bibitem[Chu et~al.(2024)Chu, Wang, and Abrantes]{chu-2024-llm_ad}
Peng Chu, Jiang Wang, and Andre Abrantes.
\newblock {LLM-AD}: Large language model based audio description system, 2024.

\bibitem[Clarke \& Lapata(2008)Clarke and Lapata]{clarke-2008-global}
James Clarke and Mirella Lapata.
\newblock Global inference for sentence compression: An integer linear programming approach.
\newblock \emph{Journal of Artificial Intelligence Research}, 31:\penalty0 399--429, 2008.
\newblock \doi{10.1613/jair.2433}.

\bibitem[Deganutti et~al.(2025)Deganutti, Hadfield, and Gilbert]{deganutti-2025-dante}
Adrienne Deganutti, Simon Hadfield, and Andrew Gilbert.
\newblock {DANTE-AD}: Dual-vision attention network for long-term audio description.
\newblock In \emph{IEEE/CVF Conference on Computer Vision and Pattern Recognition - Workshop on AI for Content Creation (AICC'25)}, 2025.

\bibitem[Fang et~al.(2025)Fang, Wu, Wu, Song, and Chan]{fang-2025-distinct_ad}
Bo~Fang, Wenhao Wu, Qiangqiang Wu, Yuxin Song, and Antoni~B. Chan.
\newblock {DistinctAD}: Distinctive audio description generation in contexts.
\newblock In \emph{Proceedings of the Computer Vision and Pattern Recognition Conference (CVPR)}, pp.\  13571--13581, June 2025.

\bibitem[Frohmann et~al.(2024)Frohmann, Sterner, Vuli{\'c}, Minixhofer, and Schedl]{frohmann-2024-segment}
Markus Frohmann, Igor Sterner, Ivan Vuli{\'c}, Benjamin Minixhofer, and Markus Schedl.
\newblock {Segment Any Text}: A universal approach for robust, efficient and adaptable sentence segmentation.
\newblock In Yaser Al-Onaizan, Mohit Bansal, and Yun-Nung Chen (eds.), \emph{Proceedings of the 2024 Conference on Empirical Methods in Natural Language Processing}, pp.\  11908--11941, Miami, Florida, USA, November 2024. Association for Computational Linguistics.
\newblock \doi{10.18653/v1/2024.emnlp-main.665}.
\newblock URL \url{https://aclanthology.org/2024.emnlp-main.665/}.

\bibitem[Fujita et~al.(2020)Fujita, Hirao, Kamigaito, Okumura, and Nagata]{fujita-2020-soda}
Soichiro Fujita, Tsutomu Hirao, Hidetaka Kamigaito, Manabu Okumura, and Masaaki Nagata.
\newblock {SODA}: Story oriented dense video captioning evaluation framework.
\newblock In \emph{Proceedings of the European Conference on Computer Vision (ECCV)}, pp.\  517--531, 2020.
\newblock \doi{10.1007/978-3-030-58539-6\_31}.

\bibitem[Garg et~al.(2024)Garg, Burns, Karagol~Ayan, Bitton, Montgomery, Onoe, Bunner, Krishna, Baldridge, and Soricut]{garg-2024-imageinwords}
Roopal Garg, Andrea Burns, Burcu Karagol~Ayan, Yonatan Bitton, Ceslee Montgomery, Yasumasa Onoe, Andrew Bunner, Ranjay Krishna, Jason~Michael Baldridge, and Radu Soricut.
\newblock {I}mage{I}n{W}ords: Unlocking hyper-detailed image descriptions.
\newblock In Yaser Al-Onaizan, Mohit Bansal, and Yun-Nung Chen (eds.), \emph{Proceedings of the 2024 Conference on Empirical Methods in Natural Language Processing}, pp.\  93--127, Miami, Florida, USA, November 2024. Association for Computational Linguistics.
\newblock \doi{10.18653/v1/2024.emnlp-main.6}.
\newblock URL \url{https://aclanthology.org/2024.emnlp-main.6/}.

\bibitem[{Gemini Team}(2026)]{gemini3.1flashlite}
{Gemini Team}.
\newblock {Gemini 3.1 Flash-Lite}: Built for intelligence at scale, March 2026.
\newblock URL \url{https://blog.google/innovation-and-ai/models-and-research/gemini-models/gemini-3-1-flash-lite/}.

\bibitem[Gillick \& Favre(2009)Gillick and Favre]{gillick-2009-scalable}
Dan Gillick and Benoit Favre.
\newblock A scalable global model for summarization.
\newblock In James Clarke and Sebastian Riedel (eds.), \emph{Proceedings of the Workshop on Integer Linear Programming for Natural Language Processing}, pp.\  10--18, Boulder, Colorado, June 2009. Association for Computational Linguistics.
\newblock URL \url{https://aclanthology.org/W09-1802/}.

\bibitem[Gillick et~al.(2008)Gillick, Favre, and Hakkani-T{\"u}r]{gillick-2008-icsi}
Dan Gillick, Benoit Favre, and Dilek Hakkani-T{\"u}r.
\newblock The {ICSI} summarization system at {TAC} 2008.
\newblock In \emph{Proceedings of the Text Analysis Conference (TAC)}, 2008.

\bibitem[Gupta et~al.(2026)Gupta, Arora, Tombari, Rohrbach, and Rohrbach]{gupta-2026-visual}
Akshita Gupta, Aditya Arora, Federico Tombari, Marcus Rohrbach, and Anna Rohrbach.
\newblock From visual cues to spoken narration: Rethinking audio description.
\newblock In \emph{Proceedings of the 2026 Conference on Empirical Methods in Natural Language Processing}. Association for Computational Linguistics, 2026.
\newblock URL \url{https://arxiv.org/abs/2609.01725}.

\bibitem[{Gurobi Optimization, LLC}(2026)]{gurobi}
{Gurobi Optimization, LLC}.
\newblock {Gurobi Optimizer Reference Manual}, 2026.
\newblock URL \url{https://www.gurobi.com}.

\bibitem[Han et~al.(2023{\natexlab{a}})Han, Bain, Nagrani, Varol, Xie, and Zisserman]{han-2023-autoad_i}
Tengda Han, Max Bain, Arsha Nagrani, G\"ul Varol, Weidi Xie, and Andrew Zisserman.
\newblock {AutoAD}: Movie description in context.
\newblock In \emph{Proceedings of the IEEE/CVF Conference on Computer Vision and Pattern Recognition (CVPR)}, pp.\  18930--18940, June 2023{\natexlab{a}}.

\bibitem[Han et~al.(2023{\natexlab{b}})Han, Bain, Nagrani, Varol, Xie, and Zisserman]{han-2023-autoad_ii}
Tengda Han, Max Bain, Arsha Nagrani, Gul Varol, Weidi Xie, and Andrew Zisserman.
\newblock {AutoAD II}: The sequel - who, when, and what in movie audio description.
\newblock In \emph{Proceedings of the IEEE/CVF International Conference on Computer Vision (ICCV)}, pp.\  13645--13655, October 2023{\natexlab{b}}.

\bibitem[Han et~al.(2024)Han, Bain, Nagrani, Varol, Xie, and Zisserman]{han-2024-autoad_iii}
Tengda Han, Max Bain, Arsha Nagrani, G\"ul Varol, Weidi Xie, and Andrew Zisserman.
\newblock {AutoAD III}: The prequel - back to the pixels.
\newblock In \emph{Proceedings of the IEEE/CVF Conference on Computer Vision and Pattern Recognition (CVPR)}, pp.\  18164--18174, June 2024.

\bibitem[Khandelwal et~al.(2025)Khandelwal, Xie, Han, Bain, Nagrani, Zisserman, Varol, and Tapaswi]{khandelwal-2025-coherent}
Eshika Khandelwal, Junyu Xie, Tengda Han, Max Bain, Arsha Nagrani, Andrew Zisserman, Gül Varol, and Makarand Tapaswi.
\newblock More than a moment: Towards coherent sequences of audio descriptions, 2025.
\newblock URL \url{https://arxiv.org/abs/2510.25440}.

\bibitem[Korbar et~al.(2024)Korbar, Huh, and Zisserman]{korbar-2024-look}
Bruno Korbar, Jaesung Huh, and Andrew Zisserman.
\newblock Look, listen and recognise: Character-aware audio-visual subtitling.
\newblock In \emph{ICASSP 2024-2024 IEEE International Conference on Acoustics, Speech and Signal Processing (ICASSP)}, pp.\  2975--2979. IEEE, 2024.

\bibitem[Lambert et~al.(2013)Lambert, Guégan, and Zhou]{lambert-2013-reordering}
Anne Lambert, Marie Guégan, and Kai Zhou.
\newblock Scene reordering in movie script alignment.
\newblock In \emph{2013 11th International Workshop on Content-Based Multimedia Indexing (CBMI)}, pp.\  213--218, 2013.
\newblock \doi{10.1109/CBMI.2013.6576585}.

\bibitem[Li et~al.(2025)Li, Padmanabhuni, Cheema, Seifi, and Fazli]{li-2025-videoa11y}
Chaoyu Li, Sid Padmanabhuni, Maryam~S Cheema, Hasti Seifi, and Pooyan Fazli.
\newblock {VideoA11y}: Method and dataset for accessible video description.
\newblock In \emph{Proceedings of the 2025 CHI Conference on Human Factors in Computing Systems}, pp.\  1--29, 2025.

\bibitem[Lin et~al.(2024)Lin, Zhang, Gao, Xia, Chen, Gao, Xie, Xiao, and Shou]{lin-2024-movie_seq}
Kevin~Qinghong Lin, Pengchuan Zhang, Difei Gao, Xide Xia, Joya Chen, Ziteng Gao, Jinheng Xie, Xuhong Xiao, and Mike~Zheng Shou.
\newblock Learning video context as interleaved multimodal sequences.
\newblock In \emph{Proceedings of the European Conference on Computer Vision (ECCV)}, pp.\  375--396, 2024.

\bibitem[Liu et~al.(2023)Liu, Li, Wu, and Lee]{liu-2023-llava}
Haotian Liu, Chunyuan Li, Qingyang Wu, and Yong~Jae Lee.
\newblock Visual instruction tuning.
\newblock In A.~Oh, T.~Naumann, A.~Globerson, K.~Saenko, M.~Hardt, and S.~Levine (eds.), \emph{Advances in Neural Information Processing Systems}, volume~36, pp.\  34892--34916. Curran Associates, Inc., 2023.

\bibitem[Madnani et~al.(2007)Madnani, Zajic, Dorr, Ayan, and Lin]{madnani-2007-multiple}
Nitin Madnani, David Zajic, Bonnie Dorr, Necip~Fazil Ayan, and Jimmy Lin.
\newblock Multiple alternative sentence compressions for automatic text summarization.
\newblock In \emph{Proceedings of the Document Understanding Conference at NLT/NAACL}, 2007.

\bibitem[Martins \& Smith(2009)Martins and Smith]{martins-2009-summarization}
Andr{\'e} Martins and Noah~A. Smith.
\newblock Summarization with a joint model for sentence extraction and compression.
\newblock In James Clarke and Sebastian Riedel (eds.), \emph{Proceedings of the Workshop on Integer Linear Programming for Natural Language Processing}, pp.\  1--9, Boulder, Colorado, June 2009. Association for Computational Linguistics.
\newblock URL \url{https://aclanthology.org/W09-1801/}.

\bibitem[McDonald(2007)]{mcdonald-2007-study}
Ryan McDonald.
\newblock A study of global inference algorithms in multi-document summarization.
\newblock In Giambattista Amati, Claudio Carpineto, and Giovanni Romano (eds.), \emph{Advances in Information Retrieval}, pp.\  557--564, Berlin, Heidelberg, 2007. Springer Berlin Heidelberg.
\newblock ISBN 978-3-540-71496-5.

\bibitem[Morris et~al.(2004)Morris, Maier, and Green]{morris-2004-wip}
Andrew~Cameron Morris, Viktoria Maier, and Phil~D Green.
\newblock From {WER} and {RIL} to {MER} and {WIL}: Improved evaluation measures for connected speech recognition.
\newblock In \emph{Proceedings of Interspeech}, pp.\  2765--2768, 2004.

\bibitem[Park et~al.(2025)Park, Ye, Lee, Ka, and Han]{park-2025-narr_ad}
Jaehyeong Park, Junchel Ye, Seungkook Lee, Hyun~W. Ka, and Dongsu Han.
\newblock {NarrAD}: Automatic generation of audio descriptions for movies with rich narrative context.
\newblock In \emph{Proceedings of the Winter Conference on Applications of Computer Vision (WACV)}, pp.\  409--419, February 2025.

\bibitem[Pavel et~al.(2020)Pavel, Reyes, and Bigham]{pavel-2020-rescribe}
Amy Pavel, Gabriel Reyes, and Jeffrey~P. Bigham.
\newblock Rescribe: Authoring and automatically editing audio descriptions.
\newblock In \emph{Proceedings of the 33rd Annual ACM Symposium on User Interface Software and Technology}, UIST '20, pp.\  747--759, New York, NY, USA, 2020. Association for Computing Machinery.
\newblock ISBN 9781450375146.
\newblock \doi{10.1145/3379337.3415864}.
\newblock URL \url{https://doi.org/10.1145/3379337.3415864}.

\bibitem[{Qwen Team}(2026{\natexlab{a}})]{qwen3.5}
{Qwen Team}.
\newblock {Qwen3.5}: Towards native multimodal agents, February 2026{\natexlab{a}}.
\newblock URL \url{https://qwen.ai/blog?id=qwen3.5}.

\bibitem[{Qwen Team}(2026{\natexlab{b}})]{qwenteam-2026-qwen35omni}
{Qwen Team}.
\newblock {Qwen3.5-Omni} technical report, 2026{\natexlab{b}}.
\newblock URL \url{https://arxiv.org/abs/2604.15804}.

\bibitem[Radford et~al.(2023)Radford, Kim, Xu, Brockman, McLeavey, and Sutskever]{radford-2023-whisper}
Alec Radford, Jong~Wook Kim, Tao Xu, Greg Brockman, Christine McLeavey, and Ilya Sutskever.
\newblock Robust speech recognition via large-scale weak supervision.
\newblock In \emph{Proceedings of the 40th International Conference on Machine Learning (ICML)}, 2023.

\bibitem[Saxena \& Keller(2024)Saxena and Keller]{saxena-2024-moviesum}
Rohit Saxena and Frank Keller.
\newblock {M}ovie{S}um: An abstractive summarization dataset for movie screenplays.
\newblock In Lun-Wei Ku, Andre Martins, and Vivek Srikumar (eds.), \emph{Findings of the Association for Computational Linguistics: ACL 2024}, pp.\  4043--4050, Bangkok, Thailand, August 2024. Association for Computational Linguistics.
\newblock \doi{10.18653/v1/2024.findings-acl.239}.
\newblock URL \url{https://aclanthology.org/2024.findings-acl.239/}.

\bibitem[Sterner et~al.(2026{\natexlab{a}})Sterner, Lapata, Lascarides, and Keller]{sterner-2026-reframed}
Igor Sterner, Mirella Lapata, Alex Lascarides, and Frank Keller.
\newblock {REFRAMED}: Towards realistic audio description generation for movies.
\newblock In \emph{Proceedings of the Conference on Language Modeling (COLM)}, October 2026{\natexlab{a}}.
\newblock URL \url{https://doi.org/10.48550/arXiv.2608.09765}.

\bibitem[Sterner et~al.(2026{\natexlab{b}})Sterner, Lascarides, and Keller]{sterner-2026-contrastive}
Igor Sterner, Alex Lascarides, and Frank Keller.
\newblock Contrastive learning with narrative twins for modeling story salience.
\newblock In Vera Demberg, Kentaro Inui, and Llu{\'i}s Marquez (eds.), \emph{Proceedings of the 19th Conference of the {E}uropean Chapter of the {A}ssociation for {C}omputational {L}inguistics (Volume 1: Long Papers)}, pp.\  1528--1550, Rabat, Morocco, March 2026{\natexlab{b}}. Association for Computational Linguistics.
\newblock ISBN 979-8-89176-380-7.
\newblock \doi{10.18653/v1/2026.eacl-long.71}.
\newblock URL \url{https://aclanthology.org/2026.eacl-long.71/}.

\bibitem[Vercauteren(2007)]{vercauteren-2007-what_when_how}
Gert Vercauteren.
\newblock Towards a european guideline for audio description.
\newblock In Jorge D{\'\i}az~Cintas, Pilar Orero, and Aline Remael (eds.), \emph{Media for All: Subtitling for the Deaf, Audio Description, and Sign Language}, pp.\  139--149. Brill, Leiden, The Netherlands, 2007.
\newblock ISBN 9789401209564.
\newblock \doi{10.1163/9789401209564\_011}.
\newblock URL \url{https://brill.com/view/book/9789401209564/B9789401209564-s011.xml}.

\bibitem[Vercauteren(2016)]{vercauteren-2016-narratological}
Gert Vercauteren.
\newblock \emph{A narratological approach to content selection in audio description}.
\newblock PhD thesis, Antwerp University, Belgium, 2016.
\newblock URL \url{https://repository.uantwerpen.be/docman/irua/4a8d3c/11347.pdf}.

\bibitem[Wang et~al.(2025)Wang, Tong, Zheng, Shen, and Wang]{wang-2025-uni_ad}
Hanlin Wang, Zhan Tong, Kecheng Zheng, Yujun Shen, and Limin Wang.
\newblock Contextual {AD} narration with interleaved multimodal sequence.
\newblock In \emph{Proceedings of the Computer Vision and Pattern Recognition Conference (CVPR)}, pp.\  8372--8383, June 2025.

\bibitem[Wang et~al.(2021)Wang, Liang, Huang, Zhang, Li, and Yu]{wang-2021-toward}
Yujia Wang, Wei Liang, Haikun Huang, Yongqi Zhang, Dingzeyu Li, and Lap-Fai Yu.
\newblock Toward automatic audio description generation for accessible videos.
\newblock In \emph{Proceedings of the 2021 CHI Conference on Human Factors in Computing Systems}, CHI '21, New York, NY, USA, 2021. Association for Computing Machinery.
\newblock ISBN 9781450380966.
\newblock \doi{10.1145/3411764.3445347}.
\newblock URL \url{https://doi.org/10.1145/3411764.3445347}.

\bibitem[Woodsend \& Lapata(2010)Woodsend and Lapata]{woodsend-2010-automatic}
Kristian Woodsend and Mirella Lapata.
\newblock Automatic generation of story highlights.
\newblock In Jan Haji{\v{c}}, Sandra Carberry, Stephen Clark, and Joakim Nivre (eds.), \emph{Proceedings of the 48th Annual Meeting of the Association for Computational Linguistics}, pp.\  565--574, Uppsala, Sweden, July 2010. Association for Computational Linguistics.
\newblock URL \url{https://aclanthology.org/P10-1058/}.

\bibitem[Woodsend \& Lapata(2012)Woodsend and Lapata]{woodsend-2012-multiple}
Kristian Woodsend and Mirella Lapata.
\newblock Multiple aspect summarization using integer linear programming.
\newblock In Jun{'}ichi Tsujii, James Henderson, and Marius Pa{\c{s}}ca (eds.), \emph{Proceedings of the 2012 Joint Conference on Empirical Methods in Natural Language Processing and Computational Natural Language Learning}, pp.\  233--243, Jeju Island, Korea, July 2012. Association for Computational Linguistics.
\newblock URL \url{https://aclanthology.org/D12-1022/}.

\bibitem[Xie et~al.(2024)Xie, Han, Bain, Nagrani, Varol, Xie, and Zisserman]{xie-2024-autoad_zero}
Junyu Xie, Tengda Han, Max Bain, Arsha Nagrani, G\"ul Varol, Weidi Xie, and Andrew Zisserman.
\newblock {AutoAD-Zero}: A training-free framework for zero-shot audio description.
\newblock In \emph{Proceedings of the Asian Conference on Computer Vision (ACCV)}, pp.\  2265--2281, December 2024.

\bibitem[Xie et~al.(2025)Xie, Han, Bain, Nagrani, Khandelwal, Varol, Xie, and Zisserman]{xie-2025-shot_by_shot}
Junyu Xie, Tengda Han, Max Bain, Arsha Nagrani, Eshika Khandelwal, G\"ul Varol, Weidi Xie, and Andrew Zisserman.
\newblock {Shot-by-Shot}: Film-grammar-aware training-free audio description generation.
\newblock In \emph{Proceedings of the IEEE/CVF International Conference on Computer Vision (ICCV)}, pp.\  16503--16513, October 2025.

\bibitem[Ye et~al.(2025)Ye, Wang, Song, Zhou, Li, and Bu]{ye-2025-focusedad}
Xiaojun Ye, Chun Wang, Yiren Song, Sheng Zhou, Liangcheng Li, and Jiajun Bu.
\newblock {FocusedAD}: Character-centric movie audio description, 2025.
\newblock URL \url{https://arxiv.org/abs/2504.12157}.

\bibitem[Zhang et~al.(2024)Zhang, Lin, Yang, Wang, Li, Lin, Liu, and Wang]{zhang-2024-mm_narator}
Chaoyi Zhang, Kevin Lin, Zhengyuan Yang, Jianfeng Wang, Linjie Li, Chung-Ching Lin, Zicheng Liu, and Lijuan Wang.
\newblock {MM-Narrator}: Narrating long-form videos with multimodal in-context learning.
\newblock In \emph{Proceedings of the IEEE/CVF Conference on Computer Vision and Pattern Recognition (CVPR)}, pp.\  13647--13657, June 2024.

\end{thebibliography}
\bibliographystyle{iclr2027_conference}
\appendix

\newpage

\section{Detailed Worked Example}\label{app:worked-example}

We use the tailor scene that opens Figure~\ref{fig:dragon-tattoo} to walk through the steps taken to prepare the inputs for and solve the AD optimization problem of Section~\ref{sec:formalization}.
The input is the video.\footnote{See { \url{rottentomatoes.com/m/the_girl_with_the_dragon_tattoo/videos/ofdvX951xl2S}}\label{foot:link}}
The output is textual description elements, each with a start time for their delivery, as shown in Figure~\ref{fig:dragon-tattoo} (page~\pageref{fig:dragon-tattoo}).
The starting point for solving the optimization problem is pre-determining the required inputs.

\begin{itemize}
\itemsep0pt
\item \textbf{Permissible narration times.}
The set of Permissible narration times are the gaps between character dialogue.
In the example, character dialogue occurs in two temporal spans: in second 10--14, and in second 58--61.
The Permissible narration times are the complement: second 0--10, 14--58, and 61--97 (the end of the video).
Within the first scene, which ends at 18s, the permissible narration times are 0--10s and 14--18s.
\item \textbf{Scene description.}
The eventualities depicted in the first scene can be expressed as a prose scene description, see the descriptions in Table~\ref{fig:worked-example}.
\item \textbf{Description elements.}
The scene description can be split up into the individual eventualities that are depicted.
Using the scene description as input, this process can be seen as a form of text segmentation.
The table shows the 15 description elements.
\item \textbf{Occurrence spans.}
Each eventuality is depicted at a time that establishes it for a sighted viewer.
Temporal grounding provides this temporal information.
This interval also permits computation of the midpoint of each occurrence span (the average of the printed numbers).
\item \textbf{Description compression.}
Each description is reformulated into $K=5$ compressed variants at 0.9, 0.8, 0.7, 0.6 and 0.5 of its word count. For example, the first element \textit{Lisbeth stands with her hands on a table in an upmarket Stockholm tailor shop.}, 14 words, could have variants \textit{Lisbeth stands with her hands on a table in an upmarket tailor shop.}, 13 words; \textit{Lisbeth leans on a table in an upmarket Stockholm tailor shop.}, 11; \textit{Lisbeth leans on a table in an upmarket tailor shop.}, 10; \textit{Lisbeth stands in an upmarket Stockholm tailor shop.}, 8; and \textit{Lisbeth stands in an upmarket tailor shop.}, 7. Each description element is accordingly transformed into $K$ compressions, resulting in $K+1$ variants.
\item \textbf{Compressed durations.}
Each compressed variant is associated with a scalar that represents the duration required for narrating it.
\item \textbf{Salience.}
Each eventuality is also associated with a salience score, representing its relevance to the story being told.
Illustrative salience scores are provided in the table.
\end{itemize}

With these pre-determined inputs, the optimizer operates over three decision variables. The final value for these decision variables also serves as the output of the optimization.

\begin{itemize}
\itemsep0pt
\item \(\mathbf{x} \in \{0,1\}^{15}\) is a binary vector, with each element representing the selection (or rejection) of each of the 15 eventualities.
\item \(\mathbf{d} \in \mathbb{R}^{15}\) is a continuous vector, with each element representing the starting delivery time for its corresponding selected eventuality.
\item \(\mathbf{y} \in \{0,1\}^{15\times (K+1)}\) is a binary matrix, with each element representing the selection of each compressed variant.
\end{itemize}

The total duration required to narrate the full scene description exceeds the time available between dialogue.
The optimization resolves this conflict by selecting the subset of the inputs that maximizes the salience-based objective.
Narrating the full scene description takes 36s at 200\,WPM, and the scene leaves 14s between dialogue: 0--10s and 14--18s.
With $\Delta_{\max}=10$s, every eventuality can be placed in either gap, so the conflict must be resolved by selection and compression rather than by eligibility.
The optimal solution selects five of the 15 eventualities.
The first gap places $e_1$ at $d_1=0$s, $e_{10}$ at $d_{10}=4.0$s and $e_{11}$ at $d_{11}=6.1$s, ending at 10.0s; the second holds $e_{14}$ at 14.1s and $e_{15}$ at 16.2s, ending at 18.0s.
In full, $e_1$, $e_{10}$ and $e_{11}$ would take 10.2s and overrun the first gap, so one must be shortened by a word; because the objective weights salience by narration time, the solver shortens the least salient of the three, selecting the first compressed variant of $e_1$ listed above ($y_{1,1}=1$), while all other selected eventualities keep their full variant.
The background details about the windows, road, mannequins and the tailor's appearance are dropped, as is the moderately salient $e_{12}$, for which no time remains.

The final decision variables are such that they maximize the objective function and meet the specified constraints.
We can read these final decision variables off to generate the resulting AD script.
For each eventuality $i$, if $x_i$ indicates selection, then we read off the delivery time $d_i$ and find the single $y_{ik}$ that indicates selection.
The textual description corresponding to $y_{ik}$, alongside its delivery start time represent the generated AD script.

\begin{table}[t]
\caption{A prose scene description for a scene in \textit{The Girl with the Dragon Tattoo} (2011). Occurrence is when each description is depicted on screen and salience is a score representing the relevance of each description to the story being told. Occ. is occurrence, and Sal. is salience. Scores are illustrative. $d_i$ is the optimal delivery start time for selected eventualities; unselected eventualities are marked --.}
\label{fig:worked-example}
\centering
\small
\begin{tabular}{lll>{\raggedright\arraybackslash}p{7cm}r}
\toprule
& \textbf{Occ.} & \textbf{Sal.} & \textbf{Description element} & $d_i$ \\
\midrule
$e_{1}$  & 0s--4s   & 0.8 & \textit{Lisbeth stands with her hands on a table in an upmarket Stockholm tailor shop.} & 0.0s \\
$e_{2}$  & 0s--4s   & 0.2 & \textit{The shop has large windows} & -- \\
$e_{3}$  & 0s--4s   & 0.1 & \textit{that look out onto a road with parked cars,} & -- \\
$e_{4}$  & 0s--4s   & 0.1 & \textit{and there are two mannequin torsos in front of the window.} & -- \\
$e_{5}$  & 0s--4s   & 0.3 & \textit{A tailor is standing behind a large table.} & -- \\
$e_{6}$  & 0s--4s   & 0.4 & \textit{He is a bald white man,} & -- \\
$e_{7}$  & 0s--4s   & 0.2 & \textit{wearing a blue shirt and a brown vest.} & -- \\
$e_{8}$  & 0s--2s   & 0.4 & \textit{The tailor drapes a garment bag across the table} & -- \\
$e_{9}$  & 2s--4s   & 0.3 & \textit{and unzips it.} & -- \\
$e_{10}$ & 4s--6s   & 0.85 & \textit{It contains a black, leather motorcycle jacket.} & 4.0s \\
$e_{11}$ & 8s--10s  & 0.9 & \textit{It is identical to Mikael's jacket in a slide that Lisbeth is holding.} & 6.1s \\
$e_{12}$ & 8s--10s  & 0.5 & \textit{The slide shows Mikael and Erika embracing.} & -- \\
$e_{13}$ & 15s--17s & 0.5 & \textit{The tailor zips the garment bag back up.} & -- \\
$e_{14}$ & 17s--18s & 0.7 & \textit{Lisbeth looks surprised by the tailor's comment,} & 14.1s \\
$e_{15}$ & 17s--18s & 0.6 & \textit{and gives him a wistful look.} & 16.2s \\
\bottomrule
\end{tabular}
\end{table}

\section{Evaluation Metrics}\label{app:metrics}

We summarize the REFRAMED metrics; see \citet{sterner-2026-reframed}
for full details.  For the dialogue-gap metrics, gaps are intervals of
at least one second between professional dialogue subtitles, and each
reference and generated description is assigned to the gap that fully
contains it, allowing a one-second collar.
SODA-M and SODA-T are defined as
\[
\mathrm{SODA\text{-}M}
=\frac{\sum_{(i, j)\in\mathcal A^\star}\mathrm{METEOR}(r_i, \hat{r}_j)}{\tfrac{1}{2}\left(|\mathcal{R}|+|\hat{\mathcal{R}}|\right)},
\quad
\mathrm{SODA\text{-}T}
=\frac{1}{|\mathcal{R}|}\sum_{(i,j)\in\mathcal A^\star}
\mathbbm{1}\!\left[|\mathrm{mid}(I_i)-\mathrm{mid}(\hat{I}_j)|<\tau\right]
\]
where $\mathcal{R}=(r_i)$ and $\hat{\mathcal{R}}=(\hat{r}_j)$ are the reference and generated description sequences, $I_i$ and $\hat{I}_j$ their time intervals, $\mathcal{A}^\star=\arg\max_{\mathcal{A}}\sum_{(i,j)\in\mathcal{A}}\mathrm{METEOR}(r_i, \hat{r}_j)$ is their optimal monotonic alignment, $\mathrm{mid}(\cdot)$ denotes time interval midpoint, and $\tau=10$.
QEval and QEval-T are defined as
\[
    \text{QEval} = \frac{1}{N} \sum_{j=1}^{N} \mathbbm{1}(\hat{a}_j = a_j) \ ; \
    \text{QEval-T} = \frac{1}{N} \sum_{j=1}^{N} \mathbbm{1}(\hat{a}_j = a_j) \cdot T_j \ ; \ T_j = \mathbbm{1}(|\mathrm{mid}(\hat{I}_j) - \mathrm{mid}(I_j)| < \tau)
\]
where $N$ is the number of questions, $a_j$ and $\hat{a}_j$ are the gold and predicted answers, and $I_j$ and $\hat{I}_j$ are the reference and predicted attribution intervals.

Following \citet{sterner-2026-reframed}, metric scores are macro-averaged over sequences\footnote{A sequence is a contiguous group of narratively related scenes. On the challenge set, sequences contain 6.8 scenes on average and there are on average 17.8 sequences per movie.} within each movie, then over movies.
In the permutation-based significance testing, we use sequences as the permutation unit.

\section{Scene descriptions}

This appendix covers the scene descriptions from which the MILP's candidate eventualities are drawn: the prompt used to generate them from video (Appendix~\ref{sec:screenplays:generation}), the character information supplied alongside the video so that characters are named correctly (Appendix~\ref{sec:characters}), and the screenplay scene descriptions used as an alternative source in our ablations (Appendix~\ref{sub:screenplays}).

\subsection{LLM-generated description prompt}
\label{sec:screenplays:generation}

\begin{tcolorbox}[colback=white, colframe=reframed-silver, sharp corners,
  boxrule=0.5mm, breakable, title=Scene Description Prompt]
\begin{Verbatim}[fontsize=\small, breaklines=true, breakanywhere=true, breaksymbolleft={}]

Your task is to generate a description of the visuals in the provided video (no audio provided). The description should be written in the style of a scene description from a screenplay, with your output as continuous narrative prose.

Adhere to all of the below.

Dos

- Write using complete sentences in the present tense and third person.
- Include only details explicitly depicted in the visuals.
- Follow the same order as the visuals.
- Write directly about the narrative world.
- Maintain objectivity rather than providing interpretations.
- Identify characters by name only if established by provided information, otherwise by describing their physical appearance.
- Vary the density of the descriptions based on the amount of visual change.

Don'ts

- Do not output anything other than the prose description.
- Do not use meta-phrasing like video or scene.
- Do not mention cinematography details like shots or camera angles.
- Do not include audio details or conversational descriptors like speaking or listening.
- Do not include repetitive details after they have been described once, such as during a conversation without new visual information.

Once you have drafted your prose description, check that it adheres to all of the above Dos and Don'ts. If it does not, revise it and check again. If it does, respond with your description.

\end{Verbatim}
\end{tcolorbox}

\subsection{Character naming}\label{sec:characters}

Characters play a central role in movie narratives and it is essential that they are named correctly in AD.
We apply two methods that support correct character naming: (1) subtitles are augmented by including the name of the character that speaks each line, (2) following \citeposs{sterner-2026-reframed} task definition, LLMs that generate descriptions from video extracts alone are also provided with the names and faces of characters mentioned in the reference AD (overcoming the otherwise ambiguity in character naming).
These methods are orthogonal in approach (one is video-based, the other text-based) and we expect that models can benefit from access to both.

Following \citet{han-2023-autoad_ii}, we apply a face recognition tool for matching detected faces against (actor headshot, character name) pairs from IMDb.
We also filter the recognized characters according to NER-identified characters for each video extract's AD (exact-match NER: $P=97.2$, $R=96.6$, $F_1=96.9$).
Against the human-labeled test set from REFRAMED, our face recognition pipeline achieves $F_1=73.7$.
Appendix~\ref{sub:character_face_detection} gives more details.
We treat the full-movie challenge set as a blind set, skipping the NER-based filtering (i.e., the reference AD transcripts are not used for any purpose except evaluation).

In terms of dialogue speaker identification, a scalable silver-standard is provided by the SDH in REFRAMED, because on occasions when the character speaking is not apparent from the visuals, the character's name is explicitly included (e.g. ``[BOND]: The name is Bond, James Bond.'')
A second useful source of information is the Speechmatics speaker diarization provided by REFRAMED, which labels each detected dialogue with a speaker tag.
We merge these two sources of information by propagating detected speaker names to all dialogue lines with the same speaker tag (Appendix~\ref{sec:identification}).
Against a gold-standard provided by the 10 `post-production' screenplays, Speechmatics speaker diarization is highly effective at separating character speakers (confusion 12.8\%, coverage 90.1\%, purity 91.6\%, outperforming all baselines, open-weight competitors and proprietary competitors we ran, Appendix Table~\ref{tab:character_diarization}).
Combined with SDH-based speaker naming, we achieve $F_1=77.9$ on the speaker identification task (Appendix~\ref{sec:identification}).

\subsubsection{Dialogue Speaker Identification}\label{sec:identification}

The task is to label each dialogue segment with the name of the character that is speaking. Our treatment of the task follows prior work \citep{korbar-2024-look}: first we rely on automatic speaker diarization that clusters dialogues with the same speaker, then we identify the character speaking a subset of dialogues, and finally we merge these two sources of information by identifying the character name associated with all dialogues in each cluster. To operationalize this, we rely on two newly available sources of information that result in a high-quality and scalable silver-standard: Speechmatics-based speaker diarization and SDH-based character naming.

\paragraph{Diarization.}
This step clusters dialogue by speaker identity. We investigate whether state-of-the-art tools can perform this task over the hours of dialogue in a movie. Our evaluation includes the open-weight models Pyannote 3.1 and Pyannote Community 1 as well as the proprietary systems Pyannote Precision 2 and Speechmatics Enhanced. Our baselines include a single speaker baseline that groups all speech into one cluster, a changing speaker baseline that assigns a unique cluster to every distinct speech segment, and a random oracle baseline that assigns each segment a random tag drawn from a fixed pool representing the true number of speakers.

Evaluation data is derived from the ten movies with best matched screenplays (see Appendix~\ref{screenplay-match}). Performance is measured using confusion and missed detection rates, which reflect the proportion of dialogue duration assigned to an incorrect speaker and the proportion of undetected dialogue respectively. False alarm rates are omitted because our evaluation restricts analysis strictly to known gold dialogue segments. We additionally report cluster purity and coverage, which reflect the proportion of each predicted cluster belonging to a single gold speaker and the proportion of each gold speaker captured by a single predicted cluster respectively.

Table~\ref{tab:character_diarization} gives results. All open-weight and proprietary systems outperform the baselines. The best open-weight system, Pyannote Community 1, achieves a confusion rate of 27.6 and a purity of 77.9. The proprietary systems perform better, and Speechmatics Enhanced achieves the lowest confusion at 15.8 and the highest purity at 89.7, alongside a coverage of 87.1 and a minimal missed rate of 0.1. REFRAMED provides manual annotation of the Speechmatics Enhanced cluster that corresponds to the audio description narrator. Removing all audio description dialogue reduces confusion further to 12.8 and increases purity to 91.6, while incurring only a marginal increase in the missed speech rate to~1.2.

\begin{table}[htbp]
\centering
\caption{Character speaker diarization performance. Confusion is the proportion of dialogue duration assigned to an incorrect speaker; Missed Rate is the proportion of undetected dialogue. Both are computed after optimal one-to-one mapping between gold and predicted speaker tags. Purity is the proportion of each predicted cluster that belongs to a single gold speaker; Coverage is the proportion of each gold speaker captured by a single predicted cluster. Baselines: Single speaker assigns all dialogue to one tag; Changing speaker assigns a new speaker to every segment; Random Oracle K assigns each segment a random tag drawn from a fixed pool of the true number of speakers. The Final row removes all segments tagged as the AD narrator, where the narrator tag(s) are manually identified and the corresponding segments then automatically discarded (note the slightly higher Missed Rate).}
\label{tab:character_diarization}
\begin{small}
\begin{tabular}{l *{4}{>{\raggedleft\arraybackslash}p{1.4cm}}}
\toprule
\textbf{System} & \textbf{Confusion ($\downarrow$)} & \textbf{Missed Rate ($\downarrow$)} & \textbf{Purity ($\uparrow$)} & \textbf{Coverage ($\uparrow$)} \\
\midrule
\rowcolor{reframed-gray}\multicolumn{5}{c}{\textit{Baselines}} \\
\midrule
Single speaker & 68.1 & 0.0 & 31.9 & 100.0 \\
Changing speaker & 95.6 & 0.0 & 100.0 & 4.4 \\
Oracle K & 89.0 & 0.0 & 33.8 & 11.7 \\
\midrule
\rowcolor{reframed-gray}\multicolumn{5}{c}{\textit{Open-weight}} \\
\midrule
Pyannote Speaker Diarization 3.1 & 31.7 & 0.6 & 78.0 & 71.0 \\
Pyannote Community 1 & 27.6 & 0.6 & 77.9 & 75.3 \\
\midrule
\rowcolor{reframed-gray}\multicolumn{5}{c}{\textit{Proprietary}} \\
\midrule
Pyannote Precision 2 & 17.6 & 0.2 & 88.5 & 83.6 \\
Speechmatics Enhanced & 15.8 & 0.1 & 89.7 & 87.1 \\
\midrule
\rowcolor{reframed-gray}\multicolumn{5}{c}{\textit{Final (AD semi-automatically removed)}} \\
\midrule
Speechmatics Enhanced & 12.8 & 1.2 & 91.6 & 90.1 \\
\bottomrule
\end{tabular}
\end{small}

\end{table}

\paragraph{Subtitle splitting}

Since our objectives are to apply the diarization to the professional subtitle provided by REFRAMED, we need to ensure that each subtitle entry corresponds to a single dialogue speaker.
Professional movie subtitles typically adhere to guidelines in this respect: if more than one speaker is being included in a single subtitle entry, each speaker dialogue line is assigned to a separate line and each line is prefixed with a hyphen.

We identify subtitle entries containing multiple lines each prefixed with a hyphen.
We attempt to split such entries into their individual lines by aligning their textual boundaries with word-level timestamps from the REFRAMED ASR transcripts.
The split time is determined by finding an exact token match for either the final word of the preceding line or the initial word of the succeeding line within the transcript (after text normalization).
Once these boundary times are identified, we assign new start and end timestamps to each extracted dialogue line. A minimum temporal gap of 0.024 seconds is enforced between the resulting segments.
We apply this to both the SDH and dialogue-only subtitles in REFRAMED.

\paragraph{Identification.}

Our speaker-identification pipeline operates in three stages.
In the first stage, we identify the speaker of the segments that are apparent from the SDH.
Our best methods rely on an LLM annotator (Olmo 3 32B Think) in order to extract the character names.
For each SDH segment, we pass it as input to the LLM alongside contextual segments spanning 30 seconds on either side of the segment's start time (using SRT formatting).
The model is instructed to return a character name only when the speaker is unambiguous (see page~\pageref{sdh-prompt} for the prompt used).
Subtitle segments formatted with hyphen prefixes on each line (multi-speaker exchanges) are split, when there exists a direct match for the first or last word of lines to be split with the REFRAMED ASR transcripts to use as the splitting time.
Entries consisting only of bracketed non-speech content or remaining hyphenated lines are excluded before inference.

The second stage resolves each free-form name against the official cast list. A cascade matching procedure considers full names, all token subsequences, and set of common nicknames, mapping a name to a cast entry only when the form is claimed by exactly one cast member.
Descriptive entries containing generic tokens such as "Man" or "Woman" are matched only by their full form, and age-prefixed entries are retained only when the base name appears separately in the cast.

The third stage transfers these labels onto speaker clusters from the ASR and speaker diarization transcripts.
For each subtitle-level label we identify the cluster whose intervals cover the largest fraction of its duration and record a vote for that pairing.
The cast entry receiving the most votes for a given cluster is taken as its label, after consulting an externally provided partial mapping from cast entries to canonical names and filling in any missing entries from the most frequent co-occurring free-form name.
Each segment then inherits the canonical label of its cluster.

\paragraph{Evaluation.}

In terms of evaluation, the name used in the best-matching screenplay may not match the full name in the IMDb cast list, and the SDH-based names may not either.
For evaluation, we resolve all names on both sides to the IMDb cast list entry using the cascade-based procedure.
A prediction is a true positive when its resolved cast entry matches the gold entry, a false positive when the entries differ, and a false negative when a gold segment is left unlabelled or mislabelled.
We report micro-averaged precision, recall, and $F_1$.

We compare four baselines against two predictive systems, each in two configurations.
The baselines (1) label every segment with the most frequent gold cast entry in the film, (2) sample uniformly from the full official cast, (3) sample uniformly from the set of cast entries in the gold labels, or (4) sample in proportion to the frequency of those entries.
The predictive systems differ in how subtitle-level names are obtained.
The first employs heuristics to identify common formatting SDH character-name cues (all-caps or title-case names preceding a colon and bracketed names at the start of a line).
The alternative system uses LLM-based identification from SDH.

Results are in Table~\ref{tab:speaker-naming}.
The sampling-based baselines perform poorly ($F_1=2.9$, $F_1=4.9$, $F_1=18.1$).
The strongest baseline labels every segment with the most frequent gold cast entry ($F_1=32.0$), unsurprising given that most movies have one or more main protagonists.
The SDH-based approach outperforms all baselines on precision (extracting-only achieves $P=89.8$ and $82.2$ vs.\ the best baseline $32.0$).
Note that precision is imperfect here due to two factors: the SDH-based labels need to be propagated onto the dialogue-only subtitles, and resolving character names to the IMDb cast list entries can be noisy.
Recall is low ($R=3.3$ and $R=4.1$; $F_1=6.4$ and $F_1=7.9$), demonstrating that only a small fraction of SDH segments explicitly cue the speaker.
The LLM identifies more characters ($R=3.3 \rightarrow 4.1$, $+0.8$) at a cost in precision ($P=89.8 \rightarrow 82.2$, $-7.6$).

Speaker-based propagation improves recall dramatically ($R=3.3\rightarrow69.3$ and $R=4.1\rightarrow74.5$).
LLM-based extraction achieves the best overall results, exceeding the extractive variant by a modest margin ($F_1=73.7 \rightarrow 77.9$, $+4.2$).

\begin{table}
    \centering
    \caption{Character speaker identification performance. The task is to identify the IMDb name of the speaker of each dialogue segment. Most frequent character assigns the modal reference character name to every segment; random IMDb character samples uniformly from the full IMDb cast; random reference character samples uniformly from character names in the reference; random-sampled reference character samples them in proportion to their reference frequency. Extracted segments only restricts predictions to segments the system named, leaving the rest unlabelled; speaker-based propagation assigns each speaker in Speechmatics diarization the system-predicted name that wins a majority vote across its segments, then labels every gold segment with the name attached to the diarised speaker it most overlaps with.}
    \label{tab:speaker-naming}
    \begin{small}
    \begin{tabular}{lrrr}
    \toprule
          & $P$ &  $R$ & $F_1$  \\ \midrule
    \rowcolor{reframed-gray}\multicolumn{4}{c}{\textit{Baselines}} \\ \midrule
        Most frequent character          &      32.0 &   32.0 & 32.0  \\
        Random IMDb character           &       2.9 &    2.9 &  2.9  \\
        Random reference character      &       4.9 &    4.9 &  4.9  \\
        Random-sampled reference character &      18.1 &   18.1 & 18.1 \\ \midrule
        \rowcolor{reframed-gray}\multicolumn{4}{c}{\textit{SDH extraction}} \\ \midrule
        SDH, extracted segments only                &      89.8 &    3.3 &  6.4 \\
        + speaker-based propagation          &      78.6 &   69.3 & 73.7 \\ \midrule
        \rowcolor{reframed-gray}\multicolumn{4}{c}{\textit{LLM prompting}} \\ \midrule
         SDH, extracted segments only    &      82.2 &    4.1 &  7.9  \\
         + speaker-based propagation   &      81.6 &   74.5 & 77.9  \\

         \bottomrule
    \end{tabular}
    \end{small}

\end{table}

\begin{tcolorbox}[colback=white, colframe=reframed-silver, sharp corners,
  boxrule=0.5mm, breakable, title=SDH Extraction Prompt]\label{sdh-prompt}
\begin{Verbatim}[fontsize=\small, breaklines=true, breakanywhere=true, breaksymbolleft={}]

You are provided an excerpt of English movie subtitles spanning roughly one minute around a target subtitle entry. Each subtitle entry has an index, a start timecode and an end timecode.

My goal is to produce high-precision speaker-name labels for a subset of the subtitle entries for a movie. A later stage will take as input all subtitle entries and the high-precision labels for a subset of them, and handle the remaining subtitle entries. Your task is to output the character name that speaks the target subtitle entry when the speaker is unambiguous. When in doubt, output "unk".

Evidence can come from the dialogue or from non-dialogue information in the subtitles. Subtitles sometimes add explicit character naming as a part of non-dialogue information if the character is not visible on-screen. Whether each character is visible is an independent factor. One character being named does not imply the next character speaking will be named. 

Rules:

1. Only commit to a character name when there exists direct evidence of the name of the speaker of the target subtitle entry.
2. Evidence that applies to another subtitle entry cannot be used as evidence for the target subtitle entry.
3. Do not attempt to guess which subtitle entries are spoken by the same character.
4. Do not infer the speaker from outside knowledge of the movie. Rely only on evidence from the provided excerpt.
5. If your output is a character name, output the character name only, omitting any other description, dialogue, etc.

INSERT_CONTEXT

Which character spoke the following target subtitle entry?

INSERT_TARGET

Your output should be a character name or "unk".

\end{Verbatim}
\end{tcolorbox}

\subsubsection{Character Names via Subtitle Identities and AD NER}\label{sub:character_names}

AD requires both identification of the relevant on-screen characters and their consistent naming (e.g. ``Commander Bond'' as opposed to ``James Bond'').
IMDb provides movie credits, but characters are frequently listed according to their full name and this may not be how they should be referred to in AD.
The earlier character names extracted from SDH can help bridge this gap.
Still, SDH names might differ from the names used in our reference AD (note that here we only consider the character names as they are referred to in the US AD, ignoring the UK version).

Our first objective is thus to obtain a consistent mapping from IMDb character entries to the name variations used in the text, and ultimately to a single definitive name for our generation systems to use.
Linking to IMDb also allows us to extract official actor headshots.
SDH character names are extracted from the approach described in Section~\ref{sec:identification}.
To extract names from the US AD reference, we apply Named Entity Recognition (NER)\footnote{\url{https://huggingface.co/dslim/bert-large-NER}}, extracting all \texttt{PER} spans.
Against the REFRAMED human annotation, the model establishes excellent performance (exact-match $P=97.2$, $R=96.6$, $F_1=96.9$).

We attempt to resolve each SDH and AD-extracted name to exactly one IMDb character name.
Our algorithm first attempts to match full names and all token subsequences.
Unmatched names are then checked against a dictionary of English nicknames\footnote{\url{https://github.com/carltonnorthern/nicknames}}. (Matching is after normalization according to \citet{radford-2023-whisper}.)
A mapping is only finalized if the name maps unambiguously to exactly one cast member.

To determine the final name for each character, we aggregate occurrence counts for each resolved name variation across the SDH and the AD, and use the name form with the highest combined frequency.
Another output is a list of characters present in the AD for each video extract.

\subsubsection{Character Face Detection}\label{sub:character_face_detection}

We address a simplified face recognition task that does not require full face tracking, but instead identifying one face for the character that is a good match.
We apply automated face recognition to the video frames and, optionally, filter the resulting noisy predictions to those characters mentioned in the reference AD (according to the NER output).
We sample video frames, detect all visible faces, and compare against the headshots using dlib-based face recognition.\footnote{\url{https://github.com/ageitgey/face_recognition}}.
If a face receives more than one match, it is assigned to the actor to which it receives highest match.

For evaluation, we use the human-labeled test set from REFRAMED ($N=230$ character faces).
A prediction is a true positive if the predicted character name matches the gold label and the predicted face crop auto-matches the reference face. 
Our pipeline achieves high recall (71.7), but without NER-based filtering lower precision (46.7; $F_1=56.6$). 
After NER-based filtering, precision improves to 83.9, at the expense of a modest drop to recall (65.7).
Overall the pipeline achieves a final $F_1=73.7$.
Recall that since we treat the full-movie challenge set as a blind set, the NER-based filtering is skipped.

\subsection{Screenplay descriptions}\label{sub:screenplays}

\paragraph{Quantifying Screenplay--Movie Match}\label{screenplay-match}

\begin{figure}[t]
    \centering
    \begin{tikzpicture}

\pgfplotsset{
    movienight axis/.style={
    axis on top,
        width=4cm, height=4cm,
        xmin=0, xmax=100, ymin=0, ymax=100,
        xtick={0,25,50,75,100}, ytick={0,25,50,75,100},
        axis lines=left,
        grid=both,
        major grid style={line width=0.5pt, dashed},
        xlabel style={font=\small},
        ylabel style={font=\small},
        yticklabel style={font=\small},
    },
    mn marker/.style={only marks, mark=*, mark size=1.6pt},
}

\begin{groupplot}[
    group style={group size=3 by 1, horizontal sep=1.8cm},
    movienight axis,
]

\nextgroupplot[
    ylabel={$\frac{H}{N_S}$}
]
    \addplot[mn marker, color=black] coordinates {(50.3817,7.9243) (40.9639,18.9285)};
    \addplot[mn marker, color=black] coordinates {(66.4179,38.8564) (87.6289,52.8775) (84.2105,67.2244) (85.9296,49.7240)};
    \addplot[mn marker, color=black] coordinates {(98.8506,86.3084) (98.5401,82.2091) (100.0000,94.6684) (100.0000,95.5243)};

\nextgroupplot[
    ylabel={$\frac{H}{N_M}$}
]
    \addplot[mn marker, color=black] coordinates {(53.5714,8.8185) (40.7407,7.9085)};
    \addplot[mn marker, color=black] coordinates {(78.2609,47.0624) (87.1795,52.7411) (89.4737,84.2246) (95.1724,49.8407)};
    \addplot[mn marker, color=black] coordinates {(95.4128,70.6634) (93.7984,90.8381) (93.3962,83.2227) (87.2807,71.2807)};

\nextgroupplot[
ylabel={WIP}]

    \addplot[mn marker, color=black] coordinates {(26.9902,0.6988) (16.6890,1.4970)};
    \addplot[mn marker, color=black] coordinates {(51.9792,18.2867) (76.3944,27.8881) (75.3463,56.6194) (81.7813,24.7828)};
    \addplot[mn marker, color=black] coordinates {(94.3161,60.9885) (92.4291,74.6772) (93.3962,78.7856) (87.2807,68.0904)};

\end{groupplot}

\end{tikzpicture}

\begin{tikzpicture}

\pgfplotsset{
    movienight axis/.style={
    axis on top,
        width=4cm, height=4cm,
        xmin=0, xmax=100, ymin=0, ymax=100,
        xtick={0,25,50,75,100}, ytick={0,25,50,75,100},
        axis lines=left,
        grid=both,
        major grid style={line width=0.5pt, dashed},
        xlabel style={font=\small},
        ylabel style={font=\small},
        yticklabel style={font=\small},
    },
    mn marker/.style={only marks, mark=*, mark size=1.6pt},
}

\begin{groupplot}[
    group style={group size=3 by 1, horizontal sep=1.8cm},
    movienight axis,
]

\nextgroupplot[
    xlabel={$\frac{|\mathcal{S}_{aligned}|}{|\mathcal{S}|}$},
    ylabel={$\frac{H}{N_S}$}
]
    \addplot[mn marker, color=black] coordinates {(51.1450,7.9243) (45.7831,18.9285)};
    \addplot[mn marker, color=black] coordinates {(62.6866,38.8564) (84.5361,52.8775) (82.7068,67.2244) (81.9095,49.7240)};
    \addplot[mn marker, color=black] coordinates {(88.5057,86.3084) (94.8905,82.2091) (95.7746,94.6684) (92.5926,95.5243)};

\nextgroupplot[
    xlabel={$\frac{|\mathcal{M}_{aligned}|}{|\mathcal{M}|}$},
    ylabel={$\frac{H}{N_M}$}
]
    \addplot[mn marker, color=black] coordinates {(57.1429,8.8185) (39.8148,7.9085)};
    \addplot[mn marker, color=black] coordinates {(69.5652,47.0624) (75.6410,52.7411) (86.8421,84.2246) (86.8966,49.8407)};
    \addplot[mn marker, color=black] coordinates {(79.8165,70.6634) (93.0233,90.8381) (87.7358,83.2227) (73.6842,71.2807)};

\nextgroupplot[xlabel={SIP}, ylabel={WIP}]

    \addplot[mn marker, color=black] coordinates {(29.2257,0.6988) (18.2285,1.4970)};
    \addplot[mn marker, color=black] coordinates {(43.6080,18.2867) (63.9440,27.8881) (71.8243,56.6194) (71.1766,24.7828)};
    \addplot[mn marker, color=black] coordinates {(70.6422,60.9885) (88.2702,74.6772) (84.0287,78.7856) (68.2261,68.0904)};

\end{groupplot}

\end{tikzpicture}
    \caption{Screenplay--movie matching scores (WIP) against a measure of screenplay--movie scene alignment (SIP, defined below) for the 10-movie challenge set. The top row shows SIP metrics derived using human-annotated scene alignments, while the bottom row uses our automatic scene alignment method. The left and middle columns decompose the metrics into their directional components (representing deletion and insertion, respectively).}
    \label{fig:wip-validation}
\end{figure}
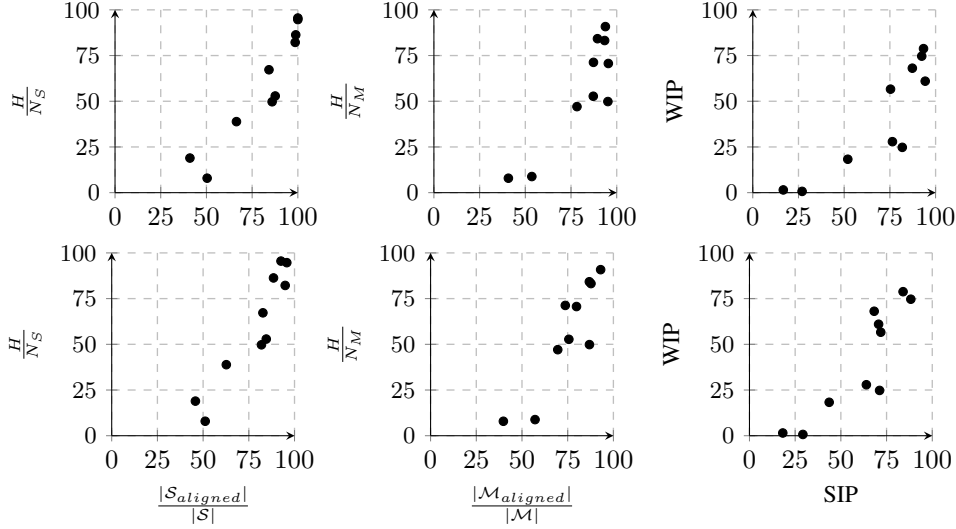

Our quantification of the degree of alignment between screenplay drafts and final film edits is based on matching the movie dialogue (from subtitles) with screenplay dialogue.
Our metric is Word Information Preserved \citep[WIP; ][]{morris-2004-wip}, a symmetric, bounded and information-theoretic metric that measures the mutual information between the two dialogue sources.
We demonstrate the effectiveness of this metric by showing that its continuous scores correlate with the manual human annotation of movie--screenplay alignment from REFRAMED (Figure~\ref{fig:wip-validation}; Spearman $\rho=0.87$, $p=0.002$ with human movie--screenplay scene alignments, and $\rho=0.83$, $p=0.005$ with automatic alignment). 
More details follow the below description on QA evaluation.

\paragraph{Description quality-assessment with QA evaluation}\label{screenplay-quality}
We evaluate the quality of LLM-generated scene descriptions using QEval on the validation set (i.e., answering AD-generated questions using the generated scene descriptions).
Reference scene descriptions extracted from the movie screenplay answer 52.9\% of the questions correctly, demonstrating their relevance for the task.
This result also shows that screenplays do not describe a substantial portion of the visual content included in audio description, a result of the flexibility afforded to movie directors and post-production editors. 
Using Gemini 3.1 Flash-Lite as an automatic LLM scene describer, we achieve 47.7\% on the same task.
Adding the automatic character faces improves performance to 48.6\%, and further adding speaker timing results in 50.6\%.
These results demonstrate that LLMs approach the QA performance of screenplay descriptions.
Results for 30 sampled movies in the REFRAMED dataset against WIP are shown in Figure~\ref{fig:screenplay-match-all}; the relationship is moderate, Spearman $\rho=0.52$, $p=0.003$, with QEval rising by about 9 points across the full WIP range. Results for the ten manually aligned challenge-set movies are shown in Figure~\ref{fig:screenplay-match-challenge}.
\begin{figure}[t]
    \centering
    \begin{minipage}[b]{0.48\textwidth}
        \centering
        \begin{tikzpicture}

\pgfplotsset{
    qeval axis/.style={
        axis on top,
        width=5cm, height=5cm,
        xmin=0, xmax=100, ymin=30, ymax=70,
        xtick={0,20,40,60,80,100},
        ytick={40,50,60},
        axis lines=left,
        grid=both,
        major grid style={line width=0.5pt, dashed, gray!50},
        xlabel style={font=\small},
        ylabel style={font=\small},
        ticklabel style={font=\small},
    },
    qeval marker/.style={only marks, mark=*, mark size=1.6pt},
}

\begin{axis}[
    qeval axis,
    xlabel={WIP},
    ylabel={QEval},
]
    \addplot[qeval marker, color=black] coordinates {(0.4251,47.2766) (1.8526,47.1270) (3.4572,44.2674) (2.2959,47.8832) (5.6584,47.7431) (7.4415,43.7118) (6.3253,52.7955) (6.9421,48.4520) (9.7767,32.1038) (9.5026,52.3296)};
    \addplot[qeval marker, color=black] coordinates {(16.9002,48.9125) (12.5293,47.7043) (23.2445,59.3889) (20.7056,51.7790) (39.2723,55.8475) (32.9953,49.8490) (42.4116,52.1712) (46.4263,47.8189) (52.6640,50.0296) (54.8997,52.4338)};
    \addplot[qeval marker, color=black] coordinates {(63.0200,53.6711) (65.4414,52.2739) (72.4278,56.7141) (67.4570,54.9163) (82.5795,51.8672) (64.8604,58.4670) (64.2033,47.2312) (66.5231,47.2074) (72.0272,60.0563) (87.1403,50.4739)};
    
    \addplot[no markers, line width=1pt, color=black, domain=0:100, samples=2] {0.089764*x + 47.121225};
\end{axis}

\end{tikzpicture}
        \caption{QEval scores using scene descriptions extracted from screenplays. Each point represents a single movie, and WIP is our measure of the degree of match between the movie's screenplay and the post-production movie.}
        \label{fig:screenplay-match-all}
    \end{minipage}%
    \hfill
    \begin{minipage}[b]{0.48\textwidth}
        \centering
        \begin{tikzpicture}

\pgfplotsset{
    qeval axis/.style={
        axis on top,
        width=5cm, height=5cm,
        xmin=0, xmax=100, ymin=30, ymax=70,
        xtick={0,20,40,60,80,100},
        ytick={40,50,60},
        axis lines=left,
        grid=both,
        major grid style={line width=0.5pt, dashed, gray!50},
        xlabel style={font=\small},
        ylabel style={font=\small},
        ticklabel style={font=\small},
    },
    qeval marker/.style={only marks, mark=*, mark size=1.6pt},
}

\begin{axis}[
    qeval axis,
    xlabel={WIP},
    ylabel={QEval},
    legend style={draw=none, fill=none, font=\small, at={(0.5,-0.22)}, anchor=north},
]
    \addplot[qeval marker, color=black] coordinates {(60.9885,56.9900) (1.4970,36.0000) (0.6988,42.6200) (74.6772,63.8300) (78.7856,49.6100) (18.2867,42.0300) (27.8881,54.3100) (56.6194,41.6400) (68.0904,55.7700) (24.7828,60.8000)};
    \addplot[no markers, line width=1pt, color=black, domain=0:100, samples=2] {0.1753*x + 43.1318};
\end{axis}

\end{tikzpicture}
        \caption{QEval scores against our measure of screenplay--movie match, WIP. The ten points represent each of the ten movies in the REFRAMED challenge set, which benefit from human-labeled screenplay alignment.}
        \label{fig:screenplay-match-challenge}
    \end{minipage}
\end{figure}

\paragraph{More on screenplay-movie scoring.} Screenplays in the MovieSum dataset \citep{saxena-2024-moviesum}, as used for REFRAMED, are from publicly available sources.  They lack metadata of the date or draft version.
The screenplays vary from early drafts to post-production versions.
Since they can deviate from the final movie, it is of practical value to quantify the degree of alignment between the screenplay version in MovieSum and the final movie.
We follow \citet{lambert-2013-reordering} by providing a continuous quantification, but unlike them stop short of classifying the resulting scores.
Our continuous quantification is based on the information-theoretic formulation of the mutual information between two sequences operationalized by \citet{morris-2004-wip}, in particular word information preserved (WIP).
Intuitively, between two sequences of words, WIP represents the probability that any given word from the first sequence is correctly matched with an identical word in the second sequence, and vice versa.
This interpretable property makes it an attractive measure and is operationalisable by comparing the sequence of screenplay dialogue words against the sequence of words spoken in the final movie dialogue.

We compute an alignment that minimizes edit distance between screenplay dialogue and movie subtitle dialogue.\footnote{Text is normalized according to \citet{radford-2023-whisper}.}
Let $H$ be the number of matched words via minimum edit distance alignment, $N_S$ be the total number of screenplay words, and $N_M$ be the total number movie subtitle words.
WIP is defined as:
\begin{equation}
    WIP = \frac{H}{N_S} \cdot \frac{H}{N_M}
\end{equation}
It multiplies match rates in both directions, and is hence a symmetric measure bounded in [0,1].
The first component $\frac{H}{N_S}$ represents the proportion of screenplay dialogue retained in the final film (penalizing deletions), while the second component $\frac{H}{N_M}$ represents the proportion of the final movie's dialogue that is also in the screenplay (penalizing insertions).

We validate the effectiveness of WIP as a continuous measure of screenplay-movie alignment by comparing against the human-annotated screenplay-movie alignment on our challenge set.
The human annotation is at the scene level, so we introduce scene information preserved (SIP) as a structural analogue to WIP. Let $|\mathcal{S}|$ and $|\mathcal{M}|$ be the total number of scenes in the screenplay and the movie.
Our analogue for a word-level \textit{hit} is for a screenplay scene to be aligned with at least one scene in the counterpart.
Let $|\mathcal{S}_{aligned}|$ and $|\mathcal{M}_{aligned}|$ be the number of scenes with at least one aligned counterpart.
SIP is formulated as:
\begin{equation}
    SIP = \frac{|\mathcal{S}_{aligned}|}{|\mathcal{S}|} \cdot \frac{|\mathcal{M}_{aligned}|}{|\mathcal{M}|}
\end{equation}
Again, these two components can be interpreted as deletion and insertion-based metrics, respectively, and (akin to WIP) SIP provides a metric bounded in [0,1] representing the degree of alignment.
(Note that, similar to WER vs. WIP, the metric would not be bounded if we used raw edit distance.)

The top three plots in Figure~\ref{fig:wip-validation} shows plots of SIP against WIP, as well as the insertion and deletion component parts against each other.
We observe a positive correlation between WIP and SIP (top right plot, Spearman $\rho=0.87$, $p=0.002$).
The bottom three plots represent results when we substitute ground-truth scene alignment with the REFRAMED automatic screenplay--movie scene alignment, where the same correlation holds (Spearman $\rho=0.83$, $p=0.005$).

\begin{figure}
    \centering
    \input{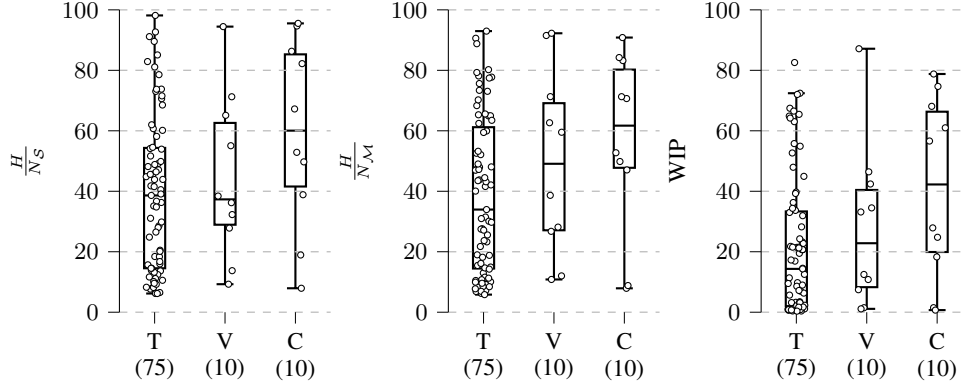}
    \caption{Distribution of WIP, and its two component parts (left two plots, representing screenplay deletion and insertion respectively), across dataset splits. T=Train, V=Validation, and C=Challenge splits; numbers in brackets represent the number of movies with aligned screenplays in each split.}
    \label{fig:reframed-wip}
\end{figure}

We compute WIP scores for all screenplays included in the REFRAMED dataset.
Figure \ref{fig:reframed-wip} shows the resulting WIP for the training and validation splits (challenge data is also provided for reference; the REFRAMED test split is not provided with screenplays).
The wide variance confirms the wide range of screenplay versions in the MovieSum dataset.

The movies with highest-WIP screenplay serve to provide a comparison source of scene descriptions for testing our AD generation methods (all from the training and validation sets).
We designate the subset of eight movies with highest WIP as a REFRAMED post-production screenplay set.\footnote{We selected the top ten, but after manual inspection, two of the movies were later excluded: \textit{The Breakfast Club (1985)}, due to its screenplay being incorrectly parsed in MovieSum (scene descriptions are inside dialogue tags), and \textit{Tár (2022)}, also due to the parsing in MovieSum (scene segmentation is not provided). Note that in both cases, the screenplays are both of high quality (in correspondence with WIP), with movie dialogue a near-perfect match against dialogue in the screenplay.}

\section{LLM processing}\label{sec:llm-processing}

The prompt below takes the video and the segmented scene description, and returns the MILP inputs for each description element: its occurrence span, its five compressed variants, and its salience score (Section~\ref{sec:setup}).

\begin{tcolorbox}[colback=white, colframe=reframed-silver, sharp corners,
  boxrule=0.5mm, breakable, title=Candidate generation prompt]
\begin{Verbatim}[fontsize=\small, breaklines=true, breakanywhere=true, breaksymbolleft={}]

Task description:

You are provided with the video for a movie scene, alongside a prose scene description. Your task is to provide temporal grounding for each extract of the scene description, and different formulations of it that could be used as audio description.

Audio description is a verbal description of key visual content in a movie. It provides blind and visually impaired individuals access to information necessary for following the story.

Output format:

Return exactly one valid JSON array and no other text. Each element represents a single extract from the prose scene description. The following illustrates the required structure for each element.

{"scene_description_extract": "SCENE_DESCRIPTION_EXTRACT", "audio_description": "DESCRIPTION", "compressed_audio_descriptions": {"0.9": "DESCRIPTION", "0.8": "DESCRIPTION", "0.7": "DESCRIPTION", "0.6": "DESCRIPTION", "0.5": "DESCRIPTION"}, "occurrence_start": "OCCURRENCE_START", "occurrence_end": "OCCURRENCE_END", "salience": SALIENCE}

- SCENE_DESCRIPTION_EXTRACT is the textual extract from the prose scene description.
- DESCRIPTION is a formulation of the same visual content as the extract that can be used as audio description.
- OCCURRENCE_START and OCCURRENCE_END indicate the temporal span in which the described visual content is established for a sighted viewer.
- SALIENCE represents the extent to which the visual content provides information necessary for following the story, relative to the other candidates in the output array. It must be a float greater than 0.0 and less than 1.0.

Rules for each element:

- SCENE_DESCRIPTION_EXTRACT must remain identical to the provided extract.
- The "audio_description" should contain a copy of the scene description extract, reformulating only as needed to form a natural audio description.
- The "compressed_audio_descriptions" dictionary must contain exactly the keys "0.9", "0.8", "0.7", "0.6", and "0.5", which refer to target compression rates of the "audio_description".
- The value for each key should target approximately that proportion of the word count of the "audio_description". Punctuation does not count as words.
- All generated descriptions must be a single sentence.
- OCCURRENCE_START and OCCURRENCE_END are timestamps with format "TIMESTAMP_FORMAT".
- OCCURRENCE_START >= 0.
- OCCURRENCE_END > OCCURRENCE_START.
- If the visual content described by an extract does not occur in the video or cannot be visually established, set OCCURRENCE_START, OCCURRENCE_END, and SALIENCE to null.

Occurrence spans for different extracts may overlap.

The following is the required output template, with the required scene description extracts included.

OUTPUT_TEMPLATE

Generate the JSON with the "audio_description", "compressed_audio_descriptions", "occurrence_start", "occurrence_end", and "salience" values filled in. Your JSON must include all elements in the template above.

\end{Verbatim}
\end{tcolorbox}

\section{LLM scheduling}\label{app:llm-scheduling}

The prompt below is used for the Qwen and Gemini scheduler baselines (Section~\ref{sec:setup}), which replace the MILP solver with an LLM given the same candidate elements and dialogue gaps.

\begin{tcolorbox}[colback=white, colframe=reframed-silver, sharp corners,
  boxrule=0.5mm, breakable, title=LLM Scheduling Prompt]
\begin{Verbatim}[fontsize=\small, breaklines=true, breakanywhere=true, breaksymbolleft={}]

Task description:

You are provided with information about a movie scene. The information is represented in two JSON arrays: (1) candidate elements containing descriptions, temporal grounding, salience scores, and narration durations; (2) the start and end times of dialogue gaps. Your task is to select a subset of the elements and assign them delivery times. The selected elements will be narrated at the assigned delivery times as audio description for the movie scene.

Audio description is a verbal description of key visual content in a movie. It provides blind and visually impaired individuals access to information necessary for following the story.

Input format:

You will receive two JSON arrays.

The first array contains candidate elements with the following structure:

{
  "element_id": "UNIQUE_ELEMENT_ID",
  "description": "DESCRIPTION_TEXT",
  "occurrence_start": NUMBER,
  "occurrence_end": NUMBER,
  "salience": NUMBER,
  "duration": NUMBER
}

The second array contains dialogue gaps with the following structure:

{
  "dialogue_gap_start": NUMBER,
  "dialogue_gap_end": NUMBER
}

All times and durations are expressed in seconds.

Output format:

Return exactly one valid JSON array and no other text. Each entry selects one element and assigns it a delivery time.

[
  {
    "element_id": "UNIQUE_ELEMENT_ID",
    "delivery_start": NUMBER
  }
]

Rules:

- The entries in your generated JSON array must contain only the keys "element_id" and "delivery_start".
- Each "element_id" must be in the JSON array of candidate elements.
- Each "element_id" may be selected at most once.
- Express "delivery_start" times in seconds.
- You may select no elements by returning [].

JSON array of elements:

ELEMENTS_JSON_ARRAY

JSON array of dialogue gaps:

DIALOGUE_GAPS_JSON_ARRAY
\end{Verbatim}
\end{tcolorbox}

\newpage

\section{System output for The Social Network (2010)}\label{sec:social-network}

The AD references shown in this section come from the REFRAMED training split, 
   where AD is automatically transcribed rather than professionally transcribed as in the 
challenge set (Section~\ref{sec:setup}). Visible errors,  for instance ``Christy \emph{sense} the
   scarf alight'' for \emph{sets}, or ``Christy \emph{drugs} \dots Eduardo into a toilet cubicle'' 
   for \emph{drags}, are artifacts of that transcription, not errors by the describer. We 
 reproduce the references verbatim, exactly as they were scored, rather than correcting them. 
 This is one reason our main results are reported on the professionally transcribed challenge 
  set, and a further reason to treat n-gram overlap against these particular references with 
   caution.
   In these examples, MILP refers to the fully automatic Qwen MILP system, and MILP (screenplay) refers to the variant that uses scene descriptions extracted from a screenplay.

\subsection{Putting Out Fires}
\begin{tcolorbox}[breakable,colback=reframed-gray!10,colframe=reframed-silver,
boxrule=0.5pt,arc=3pt,left=0pt,right=0pt,top=8pt,bottom=8pt,width=\columnwidth,fontupper=\scriptsize]
\small
\textbf{Why does your Status say "single" on your Facebook page?}\par
[+3s] \textbf{What?}\par
[+4s] \textbf{Why does your Relationship Status say "single" on your Facebook page?}\par
[+8s] \textbf{Well, I was single when I set up the page.}\par
[+10s] \textbf{And you just never bothered to change it?}\par
[+12s] \textbf{What?}\par
[+13s] \textbf{I don't know how.}\par
[+15s] \textbf{Do I look stupid to you?}\par
[+17s] \textbf{No, calm down.}\par
[+18s] \textbf{You're asking me to believe that the CFO of Facebook}\par
[+21s] \textbf{doesn't know how to change his Relationship Status on Facebook?}\par
[+23s] \textbf{It's embarrassing, so you should take it as a sign of trust that I would tell you that.}\par
[+26s] \textbf{- Go to hell. - Take it easy.}\par
[+28s] \textbf{No, you didn't change it so you could screw those Silicon Valley sluts}\par
[+31s] \textbf{every time you go out to see Mark.}\par
[+32s] \textbf{Not even remotely true, and I can promise you that the Silicon Valley sluts}\par
[+35s] \textbf{don't care what anyone's Relationship Status is on Facebook.}\par
[+38s] \textbf{Please, open your present.}\par
[+40s] \textbf{Your phone does work.}\par
[+43s] \textbf{It's Mark.}\par
[+44s] \textbf{Okay, this is gonna be tricky.}\par
[+46s] \textbf{Open your present. It's a silk scarf.}\par
[+49s] \textbf{Have you ever seen me wear a scarf?}\par
[+51s] \textbf{This'll be your first.}\par
[+53s] \textbf{Yeah.}\par
[+54s] \textbf{- You froze our account? - I did.}\par
[+55s] \textbf{- You froze the account. - I had to get your attention, Mark.}\par
[+57s] \textbf{Do you realize that you jeopardized the entire company?}\par
[+59s] \textbf{Do you realize that your actions could have destroyed}\par
[+61s] \textbf{everything I've been working on?}\par
[+62s] \textbf{We have been working on.}\par
[+63s] \textbf{Without money, the site can't function.}\par
[+65s] \textbf{Let me tell you the difference between Facebook and everybody else.}\par
[+67s] \textbf{We don't crash ever!}\par
[+68s] \textbf{If the servers are down for even a day, our entire reputation is irreversibly destroyed.}\par
[+72s] \textbf{- Look... - Users are fickle. Friendster has proved that.}\par
[+74s] \textbf{Even a few people leaving would reverberate through the entire user base.}\par
[+77s] \textbf{The users are interconnected. That is the whole point.}\par
[+79s] \textbf{College kids are online because their friends are online}\par
[+81s] \textbf{and if one domino goes, the other dominos go.}\par
[+83s] \textbf{Don't you get that?}\par
[+85s] \textbf{I am not going back to the Caribbean Night at A-E-Pi!}\par
\smallskip
{\color{reframed-gold!50!black}American AD: [+87s] \textit{Christy lights her gift on fire.}\par}
{\color{reframed-gold!50!black}British AD: [+87s] \textit{Christy sense the scarf alight.}\par}
\smallskip
{\color{reframed-red}MILP: [+87s] \textit{The woman holds a lighter}\par}
{\color{reframed-red}MILP (screenplay): [+87s] \textit{She lights the box.}\par}
\smallskip
[+88s] \textbf{Holy shit.}\par
\smallskip
{\color{reframed-gold!50!black}American AD: [+89s] \textit{She drops it into a trash can,} [+91s] \textit{then tips it onto his bed.}\par}
{\color{reframed-gold!50!black}British AD: [+91s] \textit{and drops it in a bin.}\par}
\smallskip
{\color{reframed-red}MILP: [+90s] \textit{Flames erupt from the wastebasket on the bed}\par}
{\color{reframed-red}MILP (screenplay): [+90s] \textit{He then tosses the phone down onto the bed.}\par}
\smallskip
[+92s] \textbf{What is wrong with you?}\par
[+94s] \textbf{Did you like being nobody?}\par
[+95s] \textbf{Did you like being a joke? Do you wanna go back to that?}\par
[+97s] \textbf{Hang on, hang on, hang on, hang on.}\par
[+99s] \textbf{That was the act of a child, not a businessman,}\par
[+100s] \textbf{and it certainly was not the act of a friend.}\par
[+102s] \textbf{You know how embarrassing it was for me to try to cash a check today?}\par
[+104s] \textbf{I am not going back to that life.}\par
[+106s] \textbf{- Maybe you were frustrated. - Yeah!}\par
[+108s] \textbf{- Maybe you were angry. - I was!}\par
[+111s] \textbf{But I am willing to let bygones be bygones,}\par
[+113s] \textbf{because, Wardo, I've got some good news.}\par
[+116s] \textbf{I'm sorry.}\par
[+117s] \textbf{I was angry, and maybe it was childish, but I had to get your attention.}\par
[+122s] \textbf{Wardo, I said I've got some good news.}\par
[+125s] \textbf{What is it?}\par
[+126s] \textbf{Peter Thiel just made an angel investment of half a million dollars.}\par
[+129s] \textbf{What?}\par
[+130s] \textbf{Half a million dollars. And he's setting us up in an office.}\par
[+135s] \textbf{They wanna reincorporate the company. They wanna meet you.}\par
[+137s] \textbf{They need your signature on some documents,}\par
[+139s] \textbf{so you gotta get your ass on the first flight back to San Francisco.}\par
[+141s] \textbf{I need my CFO.}\par
\smallskip
{\color{reframed-gold!50!black}American AD: [+143s] \textit{Eduardo squeezes his eyes shut,} [+144s] \textit{swallows} [+145s] \textit{and shakes his head.}\par}
{\color{reframed-gold!50!black}British AD: [+143s] \textit{Having extinguished the fire, Eduardo closes his eyes} [+146s] \textit{and shakes his head.}\par}
\smallskip
{\color{reframed-red}MILP: [+143s] \textit{He turns around to look behind him} [+145s] \textit{The woman stands in the doorway watching him closely}\par}
{\color{reframed-red}MILP (screenplay): [+144s] \textit{There is a quiet pause for a moment.}\par}
\smallskip
[+148s] \textbf{I'm on my way.}\par
[+149s] \textbf{- Wardo? - Yeah?}\par
[+151s] \textbf{We did it.}\par
\smallskip
{\color{reframed-gold!50!black}American AD: [+152s] \textit{He tearfully snaps his phone shut.}\par}
{\color{reframed-gold!50!black}British AD: [+152s] \textit{Eduardo hangs up} [+153s] \textit{and Christy appears in the doorway.} [+156s] \textit{Where are.}\par}
\smallskip
{\color{reframed-red}MILP (screenplay): [+152s] \textit{Eduardo jumps in surprise as Christy stands directly behind him.}\par}
\smallskip
[+156s] \textbf{Wardo?}\par
[+157s] \textbf{You're going back there already?}\par
[+159s] \textbf{Yes.}\par
\smallskip
\smallskip
{\color{reframed-red}MILP (screenplay): [+161s] \textit{The scene cuts away.}\par}
\smallskip
\smallskip
Sampled QA: What does Eduardo do immediately after extinguishing the fire?\par
\hfill A. calls someone \quad B. starts laughing \quad \textbf{C. closes his eyes} \quad D. leaves the room \quad E. opens the door\par
\smallskip
\end{tcolorbox}
\clearpage

\subsection{I'm Not a Bad Guy}
\begin{tcolorbox}[breakable,colback=reframed-gray!10,colframe=reframed-silver,
boxrule=0.5pt,arc=3pt,left=0pt,right=0pt,top=8pt,bottom=8pt,width=\columnwidth,fontupper=\scriptsize]
\small
\textbf{We're done for the day.}\par
[+2s] \textbf{Yeah. Yeah, I was just sitting here.}\par
\smallskip
{\color{reframed-gold!50!black}American AD: [+5s] \textit{He faces an open laptop} [+6s] \textit{and punches a few keys.}\par}
{\color{reframed-gold!50!black}British AD: [+5s] \textit{Everyone else has gone.}\par}
\smallskip
{\color{reframed-red}MILP: [+4s] \textit{Marylin enters the conference room holding a light jacket.} [+7s] \textit{Typing on laptop.}\par}
{\color{reframed-red}MILP (screenplay): [+4s] \textit{Mark types on a laptop at a table.} [+7s] \textit{She holds a coat.}\par}
\smallskip
[+8s] \textbf{What happened to Sean?}\par
[+10s] \textbf{He still owns 7\% of the company.}\par
\smallskip
{\color{reframed-gold!50!black}American AD: [+13s] \textit{He turns to face the young attorney in} [+14s] \textit{his melancholy scowl fades.}\par}
{\color{reframed-gold!50!black}British AD: [+13s] \textit{She nods.} [+14s] \textit{Mark looks up from his laptop.}\par}
\smallskip
{\color{reframed-red}MILP: [+13s] \textit{She walks to the room's middle.} [+15s] \textit{Mark halts his typing activity abruptly.}\par}
{\color{reframed-red}MILP (screenplay): [+13s] \textit{Mark looks up from his computer at Marylin.}\par}
\smallskip
[+17s] \textbf{All you had all day was that salad. Do you wanna get something to eat?}\par
[+21s] \textbf{I can't.}\par
\smallskip
{\color{reframed-gold!50!black}American AD: [+22s] \textit{Mark nods understandingly} [+24s] \textit{and briefly works his mouth.} [+25s] \textit{He turns away, shyly,} [+26s] \textit{shuts his laptop,} [+27s] \textit{then faces her again.}\par}
{\color{reframed-gold!50!black}British AD: [+24s] \textit{Mark nods and looks thoughtful.} [+26s] \textit{He opens his mouth to speak,} [+27s] \textit{then sighs and closes his laptop.}\par}
\smallskip
{\color{reframed-red}MILP: [+23s] \textit{He lifts his eyes upward.} [+24s] \textit{He makes an emphatic gesture using his right hand.} [+27s] \textit{Then he fully shuts the laptop lid downward.} [+29s] \textit{Rests hands now.}\par}
\smallskip
[+30s] \textbf{I'm not a bad guy.}\par
[+32s] \textbf{I know that.}\par
[+33s] \textbf{When there's emotional testimony, I assume 85\% of it is exaggeration.}\par
[+37s] \textbf{And the other 15?}\par
[+39s] \textbf{Perjury. Creation myths need a devil.}\par
\smallskip
{\color{reframed-gold!50!black}American AD: [+42s] \textit{Mark sighs} [+43s] \textit{and drums his laptop with his fingers.}\par}
{\color{reframed-gold!50!black}British AD: [+42s] \textit{Mark drums his fingers on his laptop.}\par}
\smallskip
{\color{reframed-red}MILP: [+42s] \textit{Marylin remains positioned at the table's head observing.} [+44s] \textit{Shifts to side.}\par}
\smallskip
[+45s] \textbf{What happens now?}\par
\smallskip
\smallskip
{\color{reframed-red}MILP: [+47s] \textit{Occupies vacant seat.}\par}
\smallskip
[+48s] \textbf{Sy and the others are having a steak on University Avenue.}\par
[+53s] \textbf{Then they'll come back up to the office,}\par
[+54s] \textbf{and start working on a settlement agreement to present to you.}\par
[+58s] \textbf{- They're gonna settle? - Oh, yeah.}\par
[+60s] \textbf{- And you're gonna have to pay a little extra. - Why?}\par
[+63s] \textbf{So that these guys sign a nondisclosure agreement.}\par
[+65s] \textbf{They say one unflattering word about you in public, you own their wife and kids.}\par
[+68s] \textbf{I invented Facebook.}\par
[+70s] \textbf{I'm talking about a jury.}\par
[+72s] \textbf{I specialize in voir dire, jury selection.}\par
[+75s] \textbf{What a jury sees when they look at a defendant.}\par
[+77s] \textbf{Clothes, hair, speaking style, likeability...}\par
[+80s] \textbf{Likeability.}\par
[+81s] \textbf{I've been licensed to practice law for all of 20 months,}\par
[+84s] \textbf{and I could get a jury to believe}\par
[+85s] \textbf{that you planted the story about Eduardo and the chicken.}\par
[+88s] \textbf{Watch what else.}\par
[+90s] \textbf{Why weren't you at Sean's sorority party that night?}\par
[+93s] \textbf{- You think I'm the one that called the police? - Doesn't matter.}\par
[+95s] \textbf{I asked the question, now everybody's thinking about it.}\par
[+97s] \textbf{You've lost your jury in the first 10 minutes.}\par
\smallskip
{\color{reframed-gold!50!black}American AD: [+99s] \textit{Mark taps his laptop.}\par}
\smallskip
{\color{reframed-red}MILP: [+99s] \textit{Positional shift made.}\par}
\smallskip
[+101s] \textbf{Farm animals.}\par
[+102s] \textbf{Yeah.}\par
\smallskip
\smallskip
{\color{reframed-red}MILP: [+103s] \textit{Bends ahead gently.}\par}
\smallskip
[+104s] \textbf{I was drunk and angry and stupid.}\par
[+107s] \textbf{- And blogging. - And blogging.}\par
[+110s] \textbf{Pay them.}\par
[+112s] \textbf{In the scheme of things,}\par
[+113s] \textbf{it's a speeding ticket.}\par
[+117s] \textbf{That's what Sy will tell you tomorrow.}\par
[+119s] \textbf{Do you think anybody would mind if I stayed and used the computer for a minute?}\par
[+122s] \textbf{I can't imagine it would be a problem.}\par
[+124s] \textbf{Thanks.}\par
[+126s] \textbf{I appreciate your help today.}\par
\smallskip
{\color{reframed-gold!50!black}American AD: [+128s] \textit{As Mark opens his laptop, the young attorney pauses and turns back.}\par}
{\color{reframed-gold!50!black}British AD: [+128s] \textit{Marylin strolls to the glass door behind him} [+131s] \textit{and turns.}\par}
\smallskip
{\color{reframed-red}MILP: [+128s] \textit{After extended silence, Marylin rises from her chair.}\par}
{\color{reframed-red}MILP (screenplay): [+128s] \textit{Marylin picks up her briefcase.}\par}
\smallskip
[+132s] \textbf{You're not an asshole, Mark.}\par
\smallskip
{\color{reframed-gold!50!black}American AD: [+134s] \textit{He faces her with a furrowed brow.}\par}
\smallskip
{\color{reframed-red}MILP: [+134s] \textit{Advances to doors.}\par}
\smallskip
[+136s] \textbf{You're just trying so hard to be.}\par
\smallskip
{\color{reframed-gold!50!black}American AD: [+137s] \textit{Marylin regards him sympathetically,} [+139s] \textit{then turns away abruptly} [+140s] \textit{and pushes out through a glass door.} [+142s] \textit{Shouldering her bag, she crosses a reception area outside.} [+146s] \textit{Mark lingers at the conference table, staring distantly.} [+149s] \textit{He shifts his gaze,} [+150s] \textit{blinks,} [+150s] \textit{then eyes his laptop screen.}\par}
{\color{reframed-gold!50!black}British AD: [+139s] \textit{He gazes at her over his shoulder.} [+141s] \textit{Marylin turns and leaves,} [+143s] \textit{slipping her handbag over one shoulder as she trots out through the adjacent office.} [+148s] \textit{Mark looks thoughtful.}\par}
\smallskip
{\color{reframed-red}MILP: [+138s] \textit{Marylin halts momentarily at the doorway entrance.} [+140s] \textit{She turns her head to observe the interior.} [+143s] \textit{Marylin departs the conference space entirely.} [+144s] \textit{Guard observed.} [+145s] \textit{Mark stays isolated at the conference table.} [+147s] \textit{Mark lifts the laptop display to reactivate it.} [+150s] \textit{He recommences data input on the keyboard.} [+152s] \textit{A fleeting screen glance precedes a pensive expression.}\par}
{\color{reframed-red}MILP (screenplay): [+138s] \textit{She walks out of the room.} [+142s] \textit{Mark settles in at his laptop.} [+146s] \textit{Mark logs onto his computer.} [+148s] \textit{Mark smiles at the screen.} [+150s] \textit{and waits for the response.}\par}
\smallskip
Sampled QA: What does Mark do after turning to face the young attorney?\par
\hfill A. waves \quad B. smiles \quad C. shrugs \quad D. frowns \quad \textbf{E. nods}\par
\smallskip
\end{tcolorbox}
\clearpage

\clearpage

\subsection{We Have Groupies}
\begin{tcolorbox}[breakable,colback=reframed-gray!10,colframe=reframed-silver,
boxrule=0.5pt,arc=3pt,left=0pt,right=0pt,top=8pt,bottom=8pt,width=\columnwidth,fontupper=\scriptsize]
\small
\textbf{So what were their names?}\par
\smallskip
\smallskip
{\color{reframed-red}MILP (screenplay): [+2s] \textit{The scene transitions to somewhere.}\par}
\smallskip
[+4s] \textbf{Their names were Christy and Alice,}\par
\smallskip
\smallskip
{\color{reframed-red}MILP: [+6s] \textit{Mark sits at a wooden table, looking serious.}\par}
{\color{reframed-red}MILP (screenplay): [+6s] \textit{Suddenly, a wooden stall door swings open.}\par}
\smallskip
[+8s] \textbf{and they wanna have drinks tonight.}\par
\smallskip
{\color{reframed-gold!50!black}American AD: [+11s] \textit{Now, in a public bathroom.}\par}
\smallskip
{\color{reframed-red}MILP: [+11s] \textit{While Eduardo leans in close, facing him.}\par}
{\color{reframed-red}MILP (screenplay): [+11s] \textit{Eduardo is forcefully shoved into the stall.}\par}
\smallskip
[+13s] \textbf{You're not supposed to be in here. This is a men's room.}\par
\smallskip
{\color{reframed-gold!50!black}American AD: [+16s] \textit{She rips her shirt off} [+18s] \textit{in Eduardo eyes.} [+19s] \textit{Her body.}\par}
{\color{reframed-gold!50!black}British AD: [+17s] \textit{Christy drugs.} [+18s] \textit{Eduardo into a toilet cubicle at a fancy club.}\par}
\smallskip
{\color{reframed-red}MILP: [+16s] \textit{The setting changes to a low angle in a bathroom.}\par}
{\color{reframed-red}MILP (screenplay): [+16s] \textit{Christy follows him in, having pushed him inside.} [+19s] \textit{She leans close to him.}\par}
\smallskip
[+21s] \textbf{Wow.}\par
\smallskip
{\color{reframed-gold!50!black}American AD: [+21s] \textit{Kissing him again, she clutches his head and neck.} [+28s] \textit{Down low on the tile floor, we glimpse Mark and Alice feet in the next stall.}\par}
{\color{reframed-gold!50!black}British AD: [+22s] \textit{She kisses him with hungry abandon.} [+24s] \textit{Hearing a noise, he pulls away.} [+36s] \textit{Mark.}\par}
\smallskip
{\color{reframed-red}MILP: [+22s] \textit{Where a woman's legs in black boots walk past a stall partition.} [+26s] \textit{Above, a young man and the woman kiss intensely.} [+28s] \textit{She holds his head.} [+29s] \textit{While he leans back against the wall.} [+32s] \textit{The view shifts underneath the stall.}\par}
{\color{reframed-red}MILP (screenplay): [+22s] \textit{She pins him against the wooden stall divider.} [+24s] \textit{Eduardo's hands slide underneath Christy's white shirt.} [+27s] \textit{His hands find her red bra just as they hear a noise.} [+30s] \textit{Someone has just entered the adjacent stall.} [+32s] \textit{Kept pinned tight.} [+34s] \textit{She reaches down to unbuckle his belt.}\par}
\smallskip
[+36s] \textbf{Oh, my God.}\par
[+38s] \textbf{I don't care.}\par
\smallskip
{\color{reframed-gold!50!black}American AD: [+39s] \textit{They kiss again} [+40s] \textit{and she unbuckles his belt,} [+42s] \textit{unzips his pants} [+43s] \textit{and reaches in.} [+44s] \textit{The view down low to the floor shows Marc's pants drop to now.} [+48s] \textit{Christy kisses her way down Eduardo's chest.} [+50s] \textit{We slowly ascend the dark wood of the stall doors,} [+54s] \textit{later standing outside the bathroom.} [+57s] \textit{A happy Eduardo glances at his friend.} [+59s] \textit{A guy approaches.}\par}
{\color{reframed-gold!50!black}British AD: [+40s] \textit{Christy feverishly unbuckles his belt} [+42s] \textit{and thrusts her hand down his boxer shorts} [+45s] \textit{in the adjacent cubicle.} [+46s] \textit{Alice tugs Mark's jeans down.} [+49s] \textit{Christy kisses Eduardo's chest} [+51s] \textit{as she sinks to her knees.} [+53s] \textit{Eduardo and Mark Stand guard outside the restroom door.} [+56s] \textit{Eduardo grins inanely at Mark, who stares into space.} [+59s] \textit{A guy approaches.}\par}
\smallskip
{\color{reframed-red}MILP: [+40s] \textit{Showing the woman standing beside the seated man among scattered shoes and bags.} [+43s] \textit{A close-up reveals hands adjusting a belt buckle before the man stands up, straightening his coat.} [+48s] \textit{Later, Mark and Eduardo stand together in a room lit by candles.} [+52s] \textit{Smiles exchanged.} [+55s] \textit{They turn their attention to a glass window.}\par}
{\color{reframed-red}MILP (screenplay): [+40s] \textit{Another noise comes from the stall next to them.} [+42s] \textit{Christy has now unzipped Eduardo's fly.} [+44s] \textit{Eduardo looks down through the gap between the stalls.} [+47s] \textit{He sees a pair of Adidas sneakers on the floor.} [+50s] \textit{Before he speaks.} [+51s] \textit{Christy pulls her shirt open to reveal the red bra.} [+54s] \textit{She places her hand inside his pants before the scene cuts.} [+57s] \textit{Mark and Eduardo stand silently outside the bathroom door.} [+60s] \textit{They exchange quiet, happy glances.}\par}
\smallskip
[+62s] \textbf{Hey, man, sorry.}\par
[+64s] \textbf{A couple girls are freshening up in there.}\par
[+67s] \textbf{Sweet.}\par
\smallskip
{\color{reframed-gold!50!black}American AD: [+68s] \textit{They share a grin} [+69s] \textit{and he goes off.} [+70s] \textit{Eduardo smile lingers.}\par}
{\color{reframed-gold!50!black}British AD: [+68s] \textit{Eduardo nods bashfully} [+70s] \textit{and grins from ear to ear.} [+71s] \textit{The guy goes.}\par}
\smallskip
{\color{reframed-red}MILP: [+68s] \textit{Where a man in a beanie grins at them.}\par}
{\color{reframed-red}MILP (screenplay): [+68s] \textit{A man approaches the door intending to use the restroom.} [+71s] \textit{Gives up leaves.}\par}
\smallskip
[+73s] \textbf{We have groupies.}\par
\smallskip
{\color{reframed-gold!50!black}American AD: [+74s] \textit{They swap a grin.} [+78s] \textit{Peering across the busy restaurant, Mark sees Erica with some friends.}\par}
{\color{reframed-gold!50!black}British AD: [+74s] \textit{Mark cracks a smile,} [+75s] \textit{and they share a little.} [+79s] \textit{Mark sees somebody in a smile fade.} [+80s] \textit{He leaves Eduardo and guard duty.}\par}
\smallskip
{\color{reframed-red}MILP: [+74s] \textit{Mark and Eduardo peer into a busy restaurant filled with diners at candlelit tables.}\par}
{\color{reframed-red}MILP (screenplay): [+74s] \textit{Taps Mark on arm.} [+75s] \textit{Mark finds himself smiling despite the awkward situation.} [+78s] \textit{Mark's expression changes as he spots something.} [+80s] \textit{Mark navigates through the crowd towards a booth.}\par}
\smallskip
[+83s] \textbf{I'll be right back.}\par
[+84s] \textbf{Mark, where you going? Mark.}\par
\smallskip
{\color{reframed-gold!50!black}American AD: [+86s] \textit{He watches uneasily as Mark approaches his ex-girlfriend's table.} [+91s] \textit{Erica.}\par}
{\color{reframed-gold!50!black}British AD: [+86s] \textit{Mark crosses the candlelit room} [+88s] \textit{and approaches a table where Erica sits with friends.}\par}
\smallskip
{\color{reframed-red}MILP: [+86s] \textit{A girl wearing a dark beret sits at a table with a group.} [+90s] \textit{Smiles while looking up.}\par}
{\color{reframed-red}MILP (screenplay): [+87s] \textit{A girl is seated at the booth.} [+89s] \textit{Although her back is turned, Mark recognizes her.}\par}
\smallskip
[+91s] \textbf{Erica?}\par
\smallskip
{\color{reframed-gold!50!black}American AD: [+91s] \textit{Erica.} [+92s] \textit{She looks up,} [+93s] \textit{then eyes him flatly.}\par}
{\color{reframed-gold!50!black}British AD: [+93s] \textit{She eyes him.}\par}
\smallskip
{\color{reframed-red}MILP: [+92s] \textit{Stops awkwardly near table shifting weight.}\par}
{\color{reframed-red}MILP (screenplay): [+93s] \textit{Surrounded by men woman.}\par}
\smallskip
[+94s] \textbf{Hi.}\par
[+95s] \textbf{I saw you from over there. I didn't know you came to this club a lot.}\par
[+98s] \textbf{- First time. - Mine, too.}\par
[+100s] \textbf{Could I talk to you alone for a second?}\par
[+103s] \textbf{I think I'm good right here.}\par
[+104s] \textbf{I just... I'd love to talk to you alone if we could just go someplace.}\par
[+108s] \textbf{Right here is fine.}\par
[+110s] \textbf{I don't know if you heard about this new website I launched.}\par
[+112s] \textbf{No.}\par
[+114s] \textbf{- The Facebook? - You called me a bitch on the Internet, Mark.}\par
[+117s] \textbf{That's why I wanted to talk to you.}\par
[+119s] \textbf{- On the Internet. - That's why I came over.}\par
[+121s] \textbf{Comparing women to farm animals.}\par
[+123s] \textbf{I didn't end up doing that.}\par
[+125s] \textbf{It didn't stop you from writing it.}\par
[+127s] \textbf{As if every thought that tumbles through your head was so clever}\par
[+129s] \textbf{it would be a crime for it not to be shared.}\par
\smallskip
{\color{reframed-gold!50!black}American AD: [+131s] \textit{Eduardo looks on.}\par}
{\color{reframed-gold!50!black}British AD: [+132s] \textit{The groupies emerge.}\par}
\smallskip
{\color{reframed-red}MILP: [+132s] \textit{Mark glances back at Eduardo.}\par}
{\color{reframed-red}MILP (screenplay): [+132s] \textit{But it hurts Mark deeply.}\par}
\smallskip
[+133s] \textbf{The Internet's not written in pencil, Mark, it's written in ink,}\par
[+136s] \textbf{and you published that Erica Albright was a bitch,}\par
[+139s] \textbf{right before you made some ignorant crack about my family's name, my bra size,}\par
[+143s] \textbf{and then rated women based on their hotness.}\par
\smallskip
\smallskip
{\color{reframed-red}MILP: [+145s] \textit{Gaze returns to girl.}\par}
\smallskip
[+146s] \textbf{- Erica, is there a problem? - No, there's no problem.}\par
\smallskip
\smallskip
{\color{reframed-red}MILP: [+149s] \textit{More emphatic gestures.}\par}
\smallskip
[+150s] \textbf{You write your snide bullshit from a dark room}\par
[+152s] \textbf{because that's what the angry do nowadays.}\par
[+155s] \textbf{I was nice to you. Don't torture me for it.}\par
[+158s] \textbf{If we could just go somewhere for a minute...}\par
[+160s] \textbf{I don't wanna be rude to my friends.}\par
[+162s] \textbf{Okay.}\par
\smallskip
{\color{reframed-gold!50!black}American AD: [+164s] \textit{He goes off.}\par}
{\color{reframed-gold!50!black}British AD: [+165s] \textit{He turns away.}\par}
\smallskip
{\color{reframed-red}MILP: [+164s] \textit{Mark eventually turns.}\par}
{\color{reframed-red}MILP (screenplay): [+164s] \textit{Turns away walks off.}\par}
\smallskip
[+166s] \textbf{Good luck with your video game.}\par
[+169s] \textbf{Hey, that was great. That was the right thing to do.}\par
[+171s] \textbf{You apologized, right?}\par
\smallskip
{\color{reframed-gold!50!black}American AD: [+172s] \textit{Mark glances aside.}\par}
{\color{reframed-gold!50!black}British AD: [+172s] \textit{Mark's eyes rove.}\par}
\smallskip
{\color{reframed-red}MILP: [+173s] \textit{And walks away down a corridor.} [+174s] \textit{By Eduardo.}\par}
{\color{reframed-red}MILP (screenplay): [+173s] \textit{He passes Eduardo and Christy, who are watching him.}\par}
\smallskip
[+175s] \textbf{We have to expand.}\par
[+177s] \textbf{Sure. Mark?}\par
\smallskip
{\color{reframed-gold!50!black}American AD: [+178s] \textit{As Mark walks out, Eduardo faces the girls.} [+181s] \textit{Something}\par}
{\color{reframed-gold!50!black}British AD: [+177s] \textit{Mark} [+178s] \textit{Mark storms out of the club.}\par}
\smallskip
{\color{reframed-red}MILP: [+178s] \textit{Two women observe scene from hall end.}\par}
{\color{reframed-red}MILP (screenplay): [+178s] \textit{Mark exits the establishment through the front door.}\par}
\smallskip
[+180s] \textbf{Is he mad about something?}\par
\smallskip
Sampled QA: Who is sitting with Erica at the table?\par
\hfill A. Mark \quad B. stranger \quad C. Eduardo \quad D. guy \quad \textbf{E. friends}\par
\end{tcolorbox}

\end{document}